\documentclass[preprint,review]{elsarticle}

\usepackage{amsmath} 
\usepackage{amssymb} 
\usepackage{units}   
\usepackage{subfig}  
\usepackage{float}   
\usepackage{graphicx}
\newtheorem{theorem}{Theorem}
\newtheorem{lemma}[theorem]{Lemma}
\newtheorem{proposition}[theorem]{Proposition}
\newtheorem{corollary}[theorem]{Corollary}
\newproof{pf}{Proof}
\DeclareGraphicsExtensions{.eps}
\floatstyle{ruled}
\newcommand{\diag}{\ensuremath{\mathrm{diag}}}
\newcommand{\MSE}[1]{\ensuremath{\mathrm{MSE}\left(#1\right)}}
\newcommand{\trace}[1]{\ensuremath{\mathrm{trace}\left( #1 \right)}}
\newcommand{\norm}[1]{\ensuremath{\left\|#1\right\|_2^2}}

\newcommand{\normzero}[1]{\ensuremath{\left\|#1\right\|_0}}
\newcommand{\real}{\ensuremath{\mathbb{R}}}

\newcommand{\supp}[1]{\ensuremath{\mathcal{#1}}}
\newcommand{\suppsize}[1]{\ensuremath{|\mathcal{#1}|}}
\newcommand{\ve}[1]{\ensuremath{\mathbf{#1}}}
\newcommand{\ves}[2]{\ensuremath{\mathbf{#1}_\supp{#2}}}
\newcommand{\mat}[1]{\ensuremath{\mathbf{#1}}}
\newcommand{\mats}[2]{\ensuremath{\mathbf{#1}_\supp{#2}}}
\newcommand{\expec}[1]{\ensuremath{E\left\{#1\right\}}}
\newcommand{\normdist}{\ensuremath{\mathcal{N}}}
\newcommand{\normdistribution}[2]{\ensuremath{\mathcal{N}\left(#1,#2\right)}}
\newcommand{\gaussian}[3]{\ensuremath{ #1 \sim  \normdist \left(#2,#3\right)} }
\newcommand{\gaussianarr}[3]{\ensuremath{ #1 & \sim & \normdist \left(#2,#3\right)} }

\journal{Applied and Computational Harmonic Analysis}

\begin{document}

\linespread{1.1}

\begin{frontmatter}
\title{On MMSE and MAP Denoising Under Sparse Representation Modeling Over a Unitary Dictionary\tnoteref{t1}}

\tnotetext[t1]{This research was supported by  the European Community's FP7-FET program, SMALL project, under grant agreement no. 225913, and by the Israel
Science Foundation (ISF) grant number 1031/08.}

\author[cstechnion]{J.S.~Turek}
\ead{javiert@cs.technion.ac.il}

\author[cstechnion]{I.~Yavneh}
\ead{irad@cs.technion.ac.il}

\author[cstechnion]{M.~Protter}
\ead{matanpr@cs.technion.ac.il}

\author[cstechnion]{M.~Elad}
\ead{elad@cs.technion.ac.il}

\address[cstechnion]{Technion, Israel Institute of Technology, Computer Science Department, Haifa 32000, Israel}

\begin{abstract}
Among the many ways to model signals, a recent approach that draws
considerable attention is sparse representation modeling. In this
model, the signal is assumed to be generated as a random linear
combination of a few atoms from a pre-specified dictionary. In
this work we analyze two Bayesian denoising algorithms -- the
Maximum-Aposteriori Probability (MAP) and the
Minimum-Mean-Squared-Error (MMSE) estimators, under the assumption
that the dictionary is unitary. It is well known that both these
estimators lead to a scalar shrinkage on the transformed
coefficients, albeit with a different response curve. In this work
we start by deriving closed-form expressions for these shrinkage
curves and then analyze their performance. Upper bounds on the MAP
and the MMSE estimation errors are derived. We tie these to the
error obtained by a so-called oracle estimator, where the support
is given, establishing a worst-case gain-factor between the
MAP/MMSE estimation errors and the oracle's performance. These
denoising algorithms are demonstrated on synthetic signals and on
true data (images).
\end{abstract}

\begin{keyword}
Sparse representations \sep MAP \sep MMSE \sep Unitary dictionary \sep Shrinkage \sep Bayesian estimation \sep Oracle
\end{keyword}
\end{frontmatter}

\linespread{1.4}
\pagebreak

\section{Introduction}
\label{sec:Intro}

A classical and long-studied subject in signal processing is
denoising. This task considers a given measurement signal $\ve{y}
\in \real^n$ obtained from a clear signal $\ve{w} \in \real^n$ by
an additive contamination of the form $\ve{y}=\ve{w}+\ve{v}$. We
shall restrict our discussion to zero mean i.i.d. Gaussian noise
vectors $\ve{v} \in \real^n$, with each entry drawn at random from
the normal distribution $\normdistribution{0}{\sigma^2}$. The
denoising goal is to recover \ve{w} from \ve{y}.

An effective denoising algorithm assumes knowledge about the noise
characteristics, like the above description, and introduces some
assumptions about the class of signals to which \ve{w} belongs,
that is, a-priori knowledge about the signal. There is a great
number of algorithms today, corresponding to a variety of signal
models. Among these, a recently emerging group of techniques
relies on sparse and redundant representations for modeling the
signals \cite{SIAM1}.

A signal \ve{w} is said to have a sparse representation over a
known dictionary, $\mat{D} \in \real^{n \times m}$, if there
exists a sparse vector $\ve{x} \in \real^m$ such that
$\ve{w}=\mat{D}\ve{x}$. The vector \ve{x} is the representation of
\ve{w}, having a number of non-zeros, $\normzero{\ve{x}}=k$, which
is much smaller than its length, $m$. Thus, \ve{x} describes how
to construct \ve{w} as a linear combination of a few columns (also
referred to as atoms) of \mat{D}. In general, the dictionary may
be redundant, containing more atoms than the signal dimension $(m
\geq n)$.

Assuming that $\ve{w} = \mat{D}\ve{x}$ with a sparse
representation \ve{x}, how can one recover \ve{w} from the noisy
measurement \ve{y}? By posing a prior probability density function
over \ve{x}, one can derive the exact  Maximum-A'posteriori
Probability (MAP) estimator for this task. This becomes a search
for the support of the sparse representation \ve{\hat x} that
maximizes the posterior probability. This problem is
computationally complex, as it generally requires an exponential
sweep over all the possible sparse supports \cite{NPHARD}.
Therefore, approximation methods are often employed, such as the
Orthogonal Matching Pursuit (OMP) \cite{OMP} and the Basis Pursuit
(BP) \cite{BP}.

While MAP estimation promotes seeking a single sparse
representation to explain the measurements, recent work has shown
that better results\footnote{In the $\ell_2$-error sense, which is
often the measure used to assess performance.} are possible using
the Minimum Mean Square Error (MMSE) estimator \cite{MMSE0, MMSE1,
MMSE2}. These works develop MMSE estimators, showing that they
lead to a weighted average of all the possible
representations that may explain the signal, with weights related
to their probabilities. Just like MAP in the general setting, this
estimation is infeasible to compute, and thus various
approximations are proposed \cite{MMSE0, MMSE1, MMSE2}.

A well known and celebrated result in signal processing is the
fact that the MAP estimator mentioned above admits a closed-form
simple formula, in the special case where the dictionary \mat{D}
is square and unitary \cite{Shrinkage1,Shrinkage2,Shrinkage3}.
This formula, known as a {\em shrinkage} operation, yields the
estimate \ve{\hat w} by applying a simple 1D operation on the
entries of the vector $\mat{D}^T \ve{y}$. The denoised signal is
then obtained by multiplication by \mat{D}. Shrinkage tends to
eliminate small entries, while leaving larger ones almost intact.

Our recent work reported in \cite{MMSE3,MMSE4} aimed to develop an
MMSE closed-form formula for the unitary case. With a specific
prior model on \ve{x}, a recursive formula for this task was
developed. Thus, at least in principle, the implications from this
work are that one need not turn to approximations, as this formula
is easily computable, leading to the exact MMSE. While such a
result is very encouraging, it does not provide a truly simple
technique of the form that MAP enjoys. Furthermore, due to its
recursive nature, this algorithm suffers from instability problems
that hinder its use for high-dimensional signals.

In the present work, we propose a modified prior model for the
sparse representation vector \ve{x}. We show that this change
leads to a simplified MMSE formula, which, just as for the MAP,
becomes a scalar shrinkage, albeit with a different response
curve. As such, this exact MMSE denoising exhibits no numerical
sensitivities as in \cite{MMSE3,MMSE4}, and thus it can operate
easily in any dimension.

The core idea that MMSE estimation for the unitary case leads to a
shrinkage algorithm has been observed before \cite{Shrinkage4,
Shrinkage5, Shrinkage6, Shrinkage7, Shrinkage8}. Here we adopt a
distinct approach in the derivation, which also gives us exact and
simple expressions for MAP and MMSE shrinkage curves, and their
expected $\ell_2$-errors. We use these as a stepping-stone towards
the development of upper bounds on the MAP and the MMSE estimation
errors.

A fundamental and key question that has attracted attention in
recent years is the proximity between practical
pursuit\footnote{Pursuit is a generic name given to algorithms
that aim to estimate \ve{x}.} results and the oracle performance.
The oracle is an estimator that knows the true support, thus
giving an ultimate result which can be used as a gold-standard for
assessing practical pursuit performance. For example, the work
reported in \cite{Danzig} shows that the Danzig Selector algorithm
is a constant (and log) factor away from the oracle result.
Similar claims for the BP, the OMP, and even the thresholding
algorithms, are made in \cite{Zvika}.

In both these papers, the analysis is deterministic and
non-Bayesian, which is different from the point of view taken in
this paper. In this work we tie the MAP and the MMSE errors for
the unitary case to the error obtained by an oracle estimator. We
establish worst-case gain-factors of the MAP and the MMSE errors
relative to the oracle error. This gives a clear ranking of these
algorithms, and states clearly their nearness to the ideal
performance.

The paper is organized as follows. In section \ref{sec:Background}
we describe the signal model we shall use throughout this work.
For completeness of the presentation, we also derive the MAP and
MMSE estimators for the general case in this section. In Section
\ref{sec:Unitary} we turn to the unitary case and present the
ideal MAP and MMSE estimators, showing how both lead to shrinkage
operations. Section \ref{sec:Bounds} is devoted to the development
of the performance behavior of the MAP and MMSE estimates, and the
upper bounds on their errors. Section \ref{sec:Experiments}
presents numerical experiments, demonstrating the proposed
algorithms in action. In Section \ref{sec:Summary} we conclude the
paper.


\section{Background}
\label{sec:Background}


\subsection{The Signal Model}
\label{sec:Model}

We consider a generative signal model that resembles the one
presented in \cite{MMSE1}. In this model, each atom has a prior
probability $P_i$ of participating in the support of each signal,
and $(1-P_i)$ of not appearing. One can think of the support
selection stage as performing biased coin-tosses of $m$ coins, with
the $\mathit{i^{th}}$ coin having a probability $P_i$ of ``heads''
and $(1-P_i)$ for ``tails''. The coins that turn up (``heads'')
constitute the support \supp{S} for this signal. Thus, the a priori
probability for any support \supp{S} is given by
\begin{equation}\label{eq:PS}
    P(\supp{S})=\prod_{i \in \supp{S}}P_i \cdot \prod_{j \notin \supp{S}}(1-P_j).
\end{equation}
It is important to note that, as opposed to the model used in
\cite{MMSE2,MMSE3,MMSE4}, here it is not possible to explicitly
prescribe the cardinality of the support, nor is it possible to
limit it (as even the empty and full supports may arise by chance). If, for
some $i$, $P_i$ equals 0, all the supports that contain element
$i$ have zero probability. Similarly, if we have $P_i=1$, then all
the supports that do not select the $i^{th}$ atom also have zero
probability. Hence, in our study we only need to consider values
$0<P_i<1$ for all $i$, and this is assumed henceforth.

We further assume that, given the support \supp{S}, the
coefficients in \ve{x} on this support are drawn as i.i.d. Gaussian
random variables\footnote{In fact, we may suggest a broader model
of the form $P(x_i) \sim \exp\{-f(x_i/\sigma_x)\}$, for an
arbitrary function $f(\cdot)$, thus keeping the model very
general. It appears that with this change one can still obtain
MMSE-shrinkage. Furthermore, one
may also study the sensitivity of MMSE/MAP shrinkage-curves
under perturbations of $f(\cdot)$, and even find the worst choice
of this function, that leads to the maximal expected error in MMSE
-- all these are left to future work, as we mainly focus here on
the Gaussian model.}
with zero mean and variance $\sigma_x^2$,
\begin{equation}
    \gaussian{\ve{x}|\supp{S}}{0}{\sigma_x^2\mat{I}_{\suppsize{S}}},
    \label{eq:distXgivenS}
\end{equation}
where $\mat{I}_{\suppsize{S}}$ is the identity matrix of size
\suppsize{S}.

We measure the vector $\ve{y} \in \real^n$, a noisy linear
combination of atoms from \mat{D} with coefficients $\ve{x} \in
\real^m$, namely, $ \ve{y}=\mat{D}\ve{x} + \ve{v}$, where the
noise \ve{v} is assumed to be white Gaussian with variance
$\sigma^2$, i.e., \gaussian{\ve{v}}{0}{\sigma^2I_n}, and the
columns of \mat{D} are normalized.

>From the model assumptions made above, it can be seen \cite{KAY}
that \ve{y} and \ve{x} are jointly Gaussians for a given support,
\begin{equation}
    \gaussian{\left.\left[
\begin{array}{c}
    \ve{y} \\ \ve{x}
\end{array}
\right]\right|\supp{S}}{\left[
\begin{array}{c}
    \ve{0} \\ \ve{0}
\end{array}
\right]}{\left[
\begin{array}{cc} 
    \mats{C}{S}             & \sigma_x^2\mats{D}{S} \\
    \sigma_x^2\mats{D}{S}^T & \sigma_x^2\mat{I}_{\suppsize{S}}
\end{array}
\right]},
    \label{eq:distXYgivenS}
\end{equation}
where
\begin{equation}
    \mats{C}{S} = \sigma_x^2\mat{D}_{\supp{S}}\mat{D}^T_{\supp{S}} + \sigma^2\mat{I}_n,
    \label{eq:matrixC}
\end{equation}and $\mat{D}_\supp{S} \in \real^{n\times\suppsize{S}}$
is comprised of the columns of the matrix \mat{D} that appear in the
support \supp{S}. Hence, the marginal p.d.f. $P(\ve{y}|\supp{S})$ is
Gaussian and it is given by
\begin{equation}
    \gaussian{\ve{y}|\supp{S}}{0}{\mats{C}{S}}.
    \label{eq:distYgivenS}
\end{equation}
Using properties of the Multivariate Gaussian p.d.f.\ (see
\cite[p. 325]{KAY}), we have that the likelihood
$P(\ve{y}|\ve{x},\supp{S})$ and the posterior p.d.f.
$P(\ve{x}|\ve{y},\supp{S})$ are also Gaussian, namely
\begin{eqnarray}
    \gaussianarr{\ve{y}|\ve{x},\supp{S}}{\mats{D}{S}\ves{x}{S}}{\sigma^2\mat{I}_n}
    \label{eq:distYgivenXS} \\
    \gaussianarr{\ve{x}|\ve{y},\supp{S}}{\frac{1}{\sigma^2}\mats{Q}{S}^{-1}
    \mats{D}{S}^T\ve{y}}{\mats{Q}{S}^{-1}},
    \label{eq:distXgivenYS}
\end{eqnarray}
where the sub-vector \ves{x}{S} is comprised of the elements of
\ve{x} whose indices are in the support \supp{S}, and
\begin{eqnarray}
    \mats{Q}{S} & = & \frac{1}{\sigma_x^2}\mat{I}_{\suppsize{S}} +
    \frac{1}{\sigma^2}\mats{D}{S}^T\mats{D}{S}.
    \label{eq:matrixQ}
\end{eqnarray}
There is a direct link between the matrices \mats{Q}{S} and
\mats{C}{S}, expressed using the matrix inversion lemma,
\begin{eqnarray}
    \mats{C}{S}^{-1} & = & \frac{1}{\sigma^2}\mat{I}_n -
    \frac{1}{\sigma^4}\mats{D}{S}\mats{Q}{S}^{-1}\mats{D}{S}^T.
    \label{eq:invMatrixC}
\end{eqnarray}


\subsection{MAP/MMSE Estimators -- The General Case}
\label{sec:General}

\subsubsection{The Oracle Estimator}
\label{subsec:GenOracle}

The first estimator we derive is the oracle. This estimator
assumes knowledge of the chosen support for \ve{x},
information that is unknown in the actual problem. Therefore it
cannot be obtained in practice. Nevertheless, it gives us a
reference performance quality to compare against. The oracle can
target the minimization of the MSE\footnote{Or MAP -- in fact, the
two are the same in this case due to the Gaussianity of
$\ve{x}|\ve{y},\supp{S}$.}. A well-known and classical result
states that the MMSE estimator is equal to the conditional mean of
the unknown, conditioned on the known parts, and thus in our case
it is $\expec{\ve{x}|\ve{y},\supp{S}}$. As the support \supp{S} is
known, we need to estimate \ves{x}{S}, the sub-vector of non-zero
entries of \ve{x}, so the estimator is given by
\begin{eqnarray}
    \ves{\hat{x}}{S}^{Oracle} = \expec{\ves{x}{S}|\ve{y},\supp{S}}
                              = \frac{1}{\sigma^2} \mats{Q}{S}^{-1}
                              \mats{D}{S}^T\ve{y}, \label{eq:Xoracle}
\end{eqnarray}
where this equality comes from the expectation of the probability
distribution in \eqref{eq:distXgivenYS}.


\subsubsection{Maximum A-Posteriori Estimator (MAP)}
\label{subsec:GenMAP}

The MAP estimator proposes an estimate \ve{\hat{x}} that maximizes
the  posterior probability. As the model mixes discrete
probabilities $P_i$ with continuous ones $P(\ve{x}|{\cal S})$, the
MAP should be carefully formulated, otherwise, the most probable
estimate would be the zero vector. Thus, we choose instead to
maximize the posterior of the support,
\begin{equation}
    \supp{S}^{MAP} = \arg\max_\supp{S} P(\supp{S}|\ve{y}),
    \label{eq:MAP-support}
\end{equation}
and only then compute the corresponding estimate
\ves{\hat{x}}{S^{MAP}}. We know from Equation
\eqref{eq:distXgivenYS}, that $\ve{x}|\ve{y},\supp{S}$ behaves as
a normal distribution, and thus the estimate \ves{\hat{x}}{MAP} is
given by the oracle in \eqref{eq:Xoracle} with the specific
support $\supp{S}^{MAP}$. Using Bayes's rule, Equation
\eqref{eq:MAP-support} leads to
\begin{equation} P(\supp{S}|\ve{y}) =
\frac{P(\ve{y}|\supp{S})P(\supp{S})}{P(\ve{y})}. \label{eq:SgivenY}
\end{equation}
Since $P(\ve{y})$ does not depend on \supp{S}, it affects this
expression only as a normalizing factor. Using the expressions of
the probabilities in the numerator that are given by Equations
\eqref{eq:distYgivenS} and \eqref{eq:PS}, respectively, we obtain
\begin{eqnarray}\label{eq:SgivenYaprox}
P(\supp{S}|\ve{y}) \propto \frac{1}{\sqrt{\det(\mats{C}{S})}}
\exp\left\{ -\frac{1}{2}\ve{y}^T\mats{C}{S}^{-1}\ve{y} \right\}
\cdot \prod_{i\in\supp{S}}P_i \cdot \prod_{j\notin\supp{S}}1-P_j
\equiv t_{\supp{S}},
\end{eqnarray}
where we have introduced the notation $t_{\supp{S}}$ for brevity of
later expressions. Returning to our MAP goal posed in Equation
\eqref{eq:MAP-support}, applying a few simple algebraic steps on the
expression for $P(\supp{S}|\ve{y})$ leads to the following penalty
function, which should be maximized with respect to the support
\supp{S},
\begin{equation}\label{eq:MAP-General}
    Val(\supp{S}) = \frac{1}{2}\norm{\frac{1}{\sigma^2}
    \mats{Q}{S}^{-1/2}\mats{D}{S}^T\ve{y}}
    -\frac{1}{2}\log\det\mats{C}{S}
    +\sum_{i\in\supp{S}}\log\left(P_i\right)
    +\sum_{j\notin\supp{S}}\log\left(1-P_j\right),
\end{equation}
over all $2^m$ possible supports. Once found, we obtain the MAP
estimation by using the oracle formula from Equation
\eqref{eq:Xoracle}, which computes \ves{\hat{x}}{\supp{S}} for
this support.


\subsubsection{Minimum Mean Square Error Estimator (MMSE)}
\label{subsec:GenMMSE}

The MMSE estimate is given by the conditional expectation,
$\expec{\ve{x}|\ve{y}}$,
\begin{equation}
\ve{\hat{x}}^{MMSE} = \expec{\ve{x}|\ve{y}}  = \int_{\ve{x}}\ve{x}P(\ve{x}|\ve{y})d\ve{x}.
\label{eq:Xmmse-0}
\end{equation}
Marginalizing the posterior probability $P(\ve{x}|\ve{y})$
over all possible supports $\supp{S}\in\Omega$, we have
\begin{equation}
    P(\ve{x}|\ve{y})  =  \sum_{\supp{S}\in\Omega}P(\ve{x}|\ve{y},\supp{S})P(\supp{S}|\ve{y}).
    \label{eq:marginalXgivenY}
\end{equation}
Plugging Equation \eqref{eq:marginalXgivenY} into Equation \eqref{eq:Xmmse-0} yields
\begin{eqnarray}
\ve{\hat{x}}^{MMSE}
& = & \sum_{\supp{S}\in\Omega} P(\supp{S}|\ve{y})
\int_{\ve{x}}\ve{x}P(\ve{x}|\ve{y},\supp{S})d\ve{x} \nonumber \\
& = & \sum_{\supp{S}\in\Omega} P(\supp{S}|\ve{y})
\expec{\ve{x}|\ve{y},\supp{S}} \nonumber \\
& = & \sum_{\supp{S}\in\Omega} P(\supp{S}|\ve{y}) \ves{\hat{x}}{S}^{Oracle}.
\label{eq:Xmmse-1}
\end{eqnarray}
Equation \eqref{eq:Xmmse-1} shows that the MMSE estimator is a
weighted average of all the ``oracle'' solutions, each with a
different support and weighted by its probability.
Finally, we substitute the expression $t_s$ developed in Equation \eqref{eq:SgivenYaprox}
into Equation \eqref{eq:Xmmse-1}, and get the formula for MMSE estimation,
\begin{equation}
\ve{\hat{x}}^{MMSE} = \frac{1}{t}\sum_{\supp{S}\in\Omega}
t_{\supp{S}} \cdot \ves{\hat{x}}{S}^{Oracle},
\label{eq:xMMSE-General}
\end{equation}
where $t=\sum_{\supp{S}\in\Omega}t_{\supp{S}}$ is the overall normalizing factor.


\subsection{Estimator Performance -- The General Case}
\label{sec:Performance}

We conclude this background section by discussing the expected
Mean-Squared-Error (MSE) induced by each of the estimators developed
above. Our goal is to obtain clear expressions for these errors,
which will later serve when we develop similar and simpler
expressions for the unitary case.

We start with the performance of the oracle estimator, as the oracle
is central to the derivation of MAP and MMSE errors. The oracle's
expected MSE is given by
\begin{eqnarray}
\expec{\left.\norm{\ves{\hat{x}}{S}^{Oracle} -
\ves{x}{S}}\right|\ve{y}} & = &
\expec{\left.\norm{\mats{Q}{S}^{-1}\frac{1}
{\sigma^2}\mats{D}{S}^T\ve{y} - \ves{x}{S}}\right|\ve{y}} \nonumber \\
& = & \expec{\left.\norm{\mats{Q}{S}^{-1}\frac{1}
{\sigma^2}\mats{D}{S}^T \left(\mats{D}{S}\ves{x}{S} + \ve{v}\right)
- \ves{x}{S}}\right| \ve{y}} = \trace{ \mats{Q}{S}^{-1}},
\label{eq:MSE-Oracle}
\end{eqnarray}
where we have used Equation \eqref{eq:matrixQ}, and the fact that
$\ve{y}=\mats{D}{S}\ves{x}{S} + \ve{v}$.

Our analysis continues with the expected error for a general
estimate $\ve{\hat{x}}$, observing that it can be written as
\begin{eqnarray}
 \expec{\left.\norm{\ve{\hat{x}} - \ve{x}}\right|\ve{y}} & = &
 \int_{\ve{x}\in\real^m}\norm{\ve{\hat{x}} -
 \ve{x}}P(\ve{x}|\ve{y})d\ve{x} \nonumber \\
 & = & \sum_{\supp{S} \in \Omega}P(\supp{S}|\ve{y})
 \int_{\ve{x}\in\real^m}\norm{\ve{\hat{x}} -
 \ve{x}}P(\ve{x}|\ve{y},\supp{S})d\ve{x} ,
 \label{eq:MSE-1}
\end{eqnarray}
where we have used the marginalization proposed in Equation
\eqref{eq:marginalXgivenY}. We add and subtract the oracle estimate
$\ves{\hat{x}}{S}^{Oracle}$ that corresponds to the support \supp{S}
into the norm term, yielding
\begin{eqnarray}
    \int_{\ve{x}\in\real^m}\norm{\ve{\hat{x}} - \ve{x}}P(\ve{x}|\ve{y},
    \supp{S})d\ve{x} & = & \int_{\ve{x}\in\real^m}\norm{\ves{\hat{x}}{S}^{Oracle}
    - \ve{x}}P(\ve{x}|\ve{y},\supp{S})d\ve{x}  \label{eq:intXgivenYS} \\
    & & + \int_{\ve{x}\in\real^m}\norm{\ve{\hat{x}} - \ves{\hat{x}}{S}^{Oracle}}
    P(\ve{x}|\ve{y},\supp{S})d\ve{x}. \nonumber
\end{eqnarray}
Note that the integral over the cross-term
$\left(\ves{\hat{x}}{S}^{Oracle} - \ve{x}
\right)^T\left(\ve{\hat{x}} - \ves{\hat{x}}{S}^{Oracle}\right)$
vanishes, since the term $\left(\ve{\hat{x}} -
\ves{\hat{x}}{S}^{Oracle}\right)$ is deterministic and can thus be
moved outside the integration, while the expression remaining
inside the integral is zero, since the oracle estimate is the
expected \ve{x} over this domain and with this support.

Continuing with Equation \eqref{eq:intXgivenYS}, the first term
represents the MSE of an oracle for a given support \supp{S}, as
derived in Equation \eqref{eq:MSE-Oracle}. In the second term, the
norm factor does not depend on the integral variable \ve{x}, and
thus it may be pulled outside the integration. The remaining part
is equal to one. Therefore,
\begin{equation}
    \int_{\ve{x}\in\real^m}\norm{\ve{\hat{x}} - \ve{x}}P(\ve{x}|\ve{y},\supp{S})d\ve{x}
    = \trace{\mats{Q}{S}^{-1}} + \norm{\ve{\hat{x}} - \ves{\hat{x}}{S}^{Oracle}}.
    \label{eq:general-intXgivenYS}
\end{equation}
Returning to the overall expected MSE as in Equation
\eqref{eq:MSE-1}, using the fact that $P(\supp{S}|\ve{y})=
t_\supp{S}/t$, as developed in Equation \eqref{eq:SgivenYaprox},
we have
\begin{equation}
    \expec{\norm{\ve{\hat{x}} - \ve{x}}} = \frac{1}{t} \sum_{\supp{S}\in\Omega}
    t_{\supp{S}} \cdot \left[ \trace{\mats{Q}{S}^{-1}} + \norm{\ve{\hat{x}}
    - \ves{\hat{x}}{S}^{Oracle}} \right].
    \label{eq:general-MSE}
\end{equation}
By plugging $\ve{\hat{x}} = \ve{\hat{x}}^{MMSE}$ into this
expression, we get the MMSE error. Note that if we minimize the
above with respect to \ve{x}, we get the MMSE estimate formula
exactly, as expected, since the MMSE is the solution that leads to
the smallest error.

Observe that \eqref{eq:general-MSE} can be written differently by
adding and subtracting $\ve{\hat{x}}^{MMSE}$ inside the norm term,
giving
\begin{eqnarray}
\expec{\left. \norm{\ve{\hat{x}} - \ve{x}}\right|} & = & \frac{1}{t}
\sum_{\supp{S}\in\Omega} t_{\supp{S}} \cdot \trace{\mats{Q}{S}^{-1}}
+ \frac{1}{t}\sum_{\supp{S}\in\Omega} t_{\supp{S}}\norm{\ve{\hat{x}}
    - \ve{\hat{x}}^{MMSE} + \ve{\hat{x}}^{MMSE} -
    \ves{\hat{x}}{S}^{Oracle}}  \nonumber \\
    & = & \frac{1}{t}
    \sum_{\supp{S}\in\Omega} t_{\supp{S}} \cdot \trace{\mats{Q}{S}^{-1}}
    +\norm{\ve{\hat{x}} - \ve{\hat{x}}^{MMSE}}
    + \frac{1}{t} \sum_{\supp{S}\in\Omega} t_{\supp{S}}
    \norm{\ve{\hat{x}}^{MMSE} - \ves{\hat{x}}{S}^{Oracle}} \nonumber \\
    & = & \norm{\ve{\hat{x}} - \ve{\hat{x}}^{MMSE}} +
    \expec{\left. \norm{\ve{\hat{x}}^{MMSE} - \ve{x}}\right|\ve{y}}.
    \label{eq:general-MSE-with-MMSE}
\end{eqnarray}
In this derivation, the cross-term $(\ve{\hat{x}} -
\ve{\hat{x}}^{MMSE})^T(\ve{\hat{x}}^{MMSE} -
\ves{\hat{x}}{S}^{Oracle})$ drops out, since in this summation the
term $(\ve{\hat{x}} - \ve{\hat{x}}^{MMSE})^T$ can be positioned
outside the summation, and then, using Equation
\eqref{eq:xMMSE-General}, it is easily shown that we are left with
an expression that equals $\ve{\hat{x}}^{MMSE} -
\ve{\hat{x}}^{MMSE} = 0$. We have then a general error formula for
any estimator, given by equation \eqref{eq:general-MSE-with-MMSE}.
In particular, this means that the error for the MAP estimate can
be calculated by
\begin{equation}
    \expec{\left.\norm{\ve{\hat{x}}^{MAP} - \ve{x}}\right|\ve{y}}
    = \norm{\ve{\hat{x}}^{MAP} -
    \ve{\hat{x}}^{MMSE}} + \expec{\left.\norm{\ve{\hat{x}}^{MMSE} -
    \ve{x}}\right|\ve{y}}.
\label{eq:general-MSE-MAP}
\end{equation}


\section{MAP \& MMSE Estimators for a Unitary Dictionary}
\label{sec:Unitary}

The derivation of MAP and MMSE for a general dictionary leads to
prohibitive computational tasks. As we shall see next, when using
unitary dictionaries, we are able to avoid these demanding
computations, and instead obtain closed-form solutions for each one
of the estimators. Furthermore, the two resulting algorithms are
very similar, both having a shrinkage structure.

While this claim about MAP and MMSE leading to shrinkage is not
new \cite{Shrinkage4,Shrinkage5,Shrinkage6,Shrinkage7,Shrinkage8},
our distinct development of the closed-form shrinkage formulae
will lead to a simple computational process for the evaluation of
the MAP and the MMSE, which will facilitate the performance
analysis derived in Section \ref{sec:Bounds}.


\subsection{The Oracle}
Just as for the general dictionary case, we start by deriving an
expression for the oracle estimation. In this case, we assume that
the dictionary \mat{D} is a unitary matrix, and thus
$\mat{D}^T\mat{D}=\mat{I}$. Moreover, it is easily seen that
$\mats{D}{S}^T\mats{D}{S}=\mat{I}_{\suppsize{S}}$, which will
simplify our expressions. We start by simplifying the matrix
$\mats{Q}{S}$ defined in \eqref{eq:matrixQ},
\begin{equation}
\mats{Q}{S} = \frac{1}{\sigma_x^2}\mat{I}_{\suppsize{S}} +
\frac{1}{\sigma^2}\mats{D}{S}^T\mats{D}{S} = \frac{\sigma_x^2+
\sigma^2}{\sigma_x^2\sigma^2}\mat{I}_{\suppsize{S}}.
\label{eq:matrixQ-Unitary}
\end{equation}
The oracle solution, as given in Equation \eqref{eq:Xoracle},
becomes
\begin{eqnarray}
    \ve{\hat{x}}^{Oracle} = \frac{1}{\sigma^2} \mats{Q}{S}^{-1}
    \mats{D}{S}^T\ve{y} = c^2 \ves{\beta}{S},
    \label{eq:Xoracle-Unitary}
\end{eqnarray}
where we have defined the constant
$c^2=\sigma_x^2/(\sigma_x^2+\sigma^2)$ and the vector
$\ves{\beta}{S}=\mats{D}{S}^T \ve{y}$. The oracle estimator has
thus been reduced to a simple matrix by vector multiplication.


\subsection{The MAP -- Unitary Case}
We turn to the MAP estimation, which requires to first find the
optimal support \supp{S} based on Equations \eqref{eq:SgivenYaprox}
and \eqref{eq:MAP-General}, and then plug it into the oracle
expression as given in Equation \eqref{eq:Xoracle} to get the
estimate.

We proceed by simplifying the expression $\det(\mats{C}{S})$ in
Equations \eqref{eq:SgivenYaprox} and \eqref{eq:MAP-General}. The
matrix \mats{C}{S} is defined in Equation \eqref{eq:matrixC} as
$\mats{C}{S} = \sigma_x^2\mats{D}{S}\mats{D}{S}^T +
\sigma^2\mat{I}_n$. Denoting by \mats{W}{S} a diagonal matrix with
ones and zeros on its main diagonal matching the
support\footnote{$(\mats{W}{S})_{ii}$ is $1$ if $i \in \supp{S}$,
and $0$ elsewhere.} \supp{S}, we obtain
\begin{eqnarray}
\det(\mats{C}{S}) & = & \det\left(\sigma_x^2\mats{D}{S}\mats{D}{S}^T +
\sigma^2\mat{I}_n\right) \nonumber \\
& = & \det\left(\mat{D}\right)\cdot  \det\left(\sigma_x^2
\mats{W}{S} + \sigma^2\mat{I}_n \right) \cdot \det\left(\mat{D}^T\right) \nonumber \\
& = & \left(\sigma_x^2+\sigma^2\right)^{\suppsize{S}}
(\sigma^2)^{n-\suppsize{S}} =
\left(1-c^2\right)^{-\suppsize{S}}\sigma^{2n}.
\label{eq:detMatrixC-Unitary}
\end{eqnarray}
Plugging this result into Equation \eqref{eq:SgivenYaprox}, and
using the relation between \mats{C}{S} and \mats{Q}{S} in Equation
\eqref{eq:invMatrixC}, yields
\begin{eqnarray}
P(\supp{S}|\ve{y}) & \propto & \exp\left\{ \frac{c^2}{2\sigma^2}
\ve{y}^T\mats{D}{S}\mats{D}{S}^T\ve{y}
 -\frac{1}{2}\log\left(1-c^2\right)^{-\suppsize{S}} \right\}
 \prod_{i\in\supp{S}}P_i \cdot \prod_{j\notin\supp{S}}1-P_j  \nonumber \\
 & \propto & \exp\left\{ \frac{c^2}{2\sigma^2}\norm{\ves{\beta}{S}}\right\}
 \prod_{i\in\supp{S}}P_i\sqrt{1-c^2} \cdot \prod_{j\notin\supp{S}}1-P_j \nonumber \\
  & \propto & \prod_{i\in\supp{S}} \exp\left\{ \frac{c^2}{2\sigma^2}
  \ve{\beta}_{i}^2\right\}P_i\sqrt{1-c^2} \cdot \prod_{j\notin\supp{S}}1-P_j.
\label{eq:SgivenY-Unitary-1}
\end{eqnarray}
Taking into account that $0<P_i<1$, we can rewrite this expression
as
\begin{eqnarray} P(\supp{S}|\ve{y}) & \propto &
\prod_{i\in\supp{S}} \exp\left\{
\frac{c^2}{2\sigma^2}\ve{\beta}_{i}^2\right\}\frac{P_i}{1-P_i}
\sqrt{1-c^2} \cdot \prod_{j=1}^n 1-P_j \nonumber \\
 & \propto & \prod_{i\in\supp{S}} \exp\left\{\frac{c^2}{2\sigma^2}
 \ve{\beta}_{i}^2\right\} \frac{P_i}{1-P_i}\sqrt{1-c^2} =
 \prod_{i\in\supp{S}} q_i, \label{eq:SgivenY-Unitary-2}
\end{eqnarray}
where we have defined
\begin{eqnarray}\label{eq:def_qi}
q_i = \exp\left\{\frac{c^2}{2\sigma^2}\ve{\beta}_{i}^2\right\} \frac{P_i}{1-P_i}\sqrt{1-c^2}.
\end{eqnarray}
We further define $g_i=q_i/(1+q_i)$ (which implies that
$q_i=g_i/(1-g_i)$), and substitute this into Equation
\eqref{eq:SgivenY-Unitary-2}. Adding now the necessary
normalization factor we get
\begin{eqnarray}
P(\supp{S}|\ve{y}) & = & \left(\sum_{\supp{S}^{*}\in\Omega}
\prod_{i\in \supp{S}^{*}}q_{i}\right)^{-1}\prod_{i\in \supp{S}}q_{i} \nonumber \\
 & = & \left(\sum_{\supp{S}^{*}\in\Omega}\prod_{i\in
 \supp{S}^{*}}\frac{g_{i}}{1-g_{i}}\right)^{-1}\prod_{i\in \supp{S}}
 \frac{g_{i}}{1-g_{i}} \nonumber \\
 & = & \left(\frac{\sum_{\supp{S}^{*}\in\Omega}\prod_{i\in
 \supp{S}^{*}}g_i\prod_{j\notin \supp{S}^{*}}\left(1-g_j\right)}
 {\prod_{k=1}^n\left(1-g_k\right)}\right)^{-1}\frac{\prod_{i\in
 \supp{S}}g_i\prod_{j\notin \supp{S}}\left(1-g_j\right)}
 {\prod_{k=1}^n\left(1-g_k\right)} \nonumber \\
 & = & \left(\sum_{\supp{S}^{*}\in\Omega}\prod_{i\in
 \supp{S}^{*}}g_{i}\prod_{j\notin \supp{S}^{*}}\left(1-g_{j}\right)\right)^{-1}
 \prod_{i\in \supp{S}}g_{i}\prod_{j\notin \supp{S}}\left(1-g_{j}\right).
\label{eq:SgivenY-Unitary-3}
\end{eqnarray}
The following observation will facilitate a further simplification
of this expression:
\begin{proposition}
Let $\Omega$ be the set of all possible subsets of $n$ indices,
and let $g_i$ be values associated with each index, such that $0
\leq g_i \leq 1$. Then,
\begin{equation}
\sum_{\supp{S} \in \Omega} \prod_{i \in \supp{S}} g_i \cdot
\prod_{j \notin \supp{S}} \left(1 - g_j\right) = 1.
\end{equation}
\label{prop1}
\end{proposition}

\begin{pf}
Consider the following experiment: a set of
$n$ independent coins are tossed, with the $i^{th}$ coin having a
probability $g_i$ for ``heads'' and $\left( 1 - g_i \right)$ for
``tails''. The probability of a specific set of $\supp{S}$ coins
turning up ``heads'' (and the rest turning up ``tails'') is
$\prod_{i \in \supp{S}} g_i \cdot \prod_{j \notin \supp{S}} \left(1
- g_j \right)$. For any one toss of the $n$ coins, exactly one of
these combinations will be the outcome. Therefore, the sum of these
probabilities over all the combinations must be $1$. \hfill $\Box$
\end{pf}

Using this proposition, the normalization term in Equation
\eqref{eq:SgivenY-Unitary-3} vanishes, as it is equal to 1 ($0<g_i
\leq 1$ since $g_i=\frac{q_i}{1+q_i}$ and $q_i \geq 0$ for every
$i$). We therefore obtain
\begin{equation}
P(\supp{S}|\ve{y})  = \prod_{i\in \supp{S}}g_{i}\prod_{j\notin \supp{S}}\left(1-g_{j}\right).
\label{eq:SgivenY-Unitary-4}
\end{equation}

The optimization task \eqref{eq:MAP-support} can now be written as
\begin{eqnarray}
\supp{S^{MAP}} & = & \arg\max_{\supp{S}\in\Omega} ~~
\prod_{i\in \supp{S}}g_{i}\prod_{j\notin \supp{S}}\left(1-g_{j}\right) \nonumber \\
& = & \arg\max_{\supp{S}\in\Omega} ~~
\prod_{i\in \supp{S}}\frac{q_{i}}{1+q_i}\prod_{j\notin \supp{S}}\left(1-\frac{q_{j}}{1+q_j}\right) \nonumber \\
& = & \arg\max_{\supp{S}\in\Omega} ~~ \frac{\prod_{i\in
\supp{S}}q_{i}\prod_{j\notin \supp{S}}1}{\prod_{k=1}^{n}(1+q_k)}=
\arg\max_{\supp{S}\in\Omega} ~~ \prod_{i\in
\supp{S}}\exp\left\{\frac{c^2}{2\sigma^2}\ve{\beta}_{i}^2\right\}
\frac{P_i}{1-P_i}\sqrt{1-c^2}.
 \label{eq:MAP-Unitary}
\end{eqnarray}
Interpreting this expression, we see that every element in the
support influences the penalty in one of two ways:
\begin{itemize}
\item {\bf If it is part of the support}: Multiply the expression
by $\sqrt{1-c^2} \frac{P_i}{1-P_i}\exp \left\{\frac{c^2}{2\sigma^2}
\ve{\beta}_i^2\right\}$, or

\item {\bf If it is not in the support}: Multiply the expression
by $1$.
\end{itemize}

\noindent As we aim to maximize the expression in Equation
\eqref{eq:MAP-Unitary}, the support will contain all the elements
$i$ such that $\sqrt{1-c^2} \cdot \frac{P_i}{1-P_i}\cdot\exp \left\{\frac{c^2}{2\sigma^2}
\ve{\beta}_i^2\right\} > 1$. (In the
case that no such element exists, the support should be empty
and the solution is therefore $\ve{\hat{x}}^{MAP}=\ve{0}$.) Once
these elements are found, all we have to do is to multiply their
value $\beta_i$ by $c^2$ and this is the MAP estimate.

Stated differently, this means that after computing the
transformed vector $\ve{\beta} = \mat{D}^T \ve{y}$, we test each
of its entries, and set the MAP estimate for the $i^{th}$ entry to be
\begin{eqnarray}\label{eq:MAPshrinkage}
    {\hat x}_i^{MAP} = \psi_{MAP}(\beta_i) = \left\{
    \begin{array}{cc}
        c^2\beta_i & |\beta_i| > \frac{\sqrt{2}\sigma}{c}
        \sqrt{\log\left(\frac{1-P_i}{\sqrt{1-c^2}P_i} \right)}\\
        0 & \mbox{otherwise}
    \end{array}
\right. .
\end{eqnarray}
This is the shrinkage algorithm mentioned earlier -- each entry is
handled independently of the others, passing through a scalar
shrinkage curve that nulls small entries and keeps large ones
intact (up to the multiplication by $c^2$). There is no trace of
the exhaustive and combinatorial search that characterizes MAP in
the general case, and this simple algorithm yields the exact MAP
estimation.


\subsection{The MMSE -- The Unitary Case}

Equation \eqref{eq:xMMSE-General} shows the presence of the oracle
in the MMSE estimation. Similarly to MAP, we make use of the
unitary oracle estimate in Equation \eqref{eq:Xoracle-Unitary}.
Note that $\ves{\beta}{S}$ may be written as
\begin{equation}
\ves{\beta}{S} = \sum_{k=1}^n\mats{I}{S}(k)\beta_k\ve{e}_k,
\label{eq:betaS}
\end{equation}
where $\ve{e}_k$ is the $k^{th}$ vector in the canonical basis,
and $\mats{I}{S}(k)$ is an indicator function ($\mats{I}{S}(k)=1$
if $k\in\supp{S}$, and zero otherwise). While this may seem like a
cumbersome change, it will prove valuable in later derivations.
Starting from Equation \eqref{eq:xMMSE-General}, substituting the
expression developed for $P(\supp{S}|\ve{y})$ in Equation
\eqref{eq:SgivenY-Unitary-4} into Equation
\eqref{eq:xMMSE-General}, and using Equation \eqref{eq:betaS}, we
obtain the following expression for the unitary MMSE estimator,
\begin{eqnarray}\label{eq:xMMSE-Unitary}
\ve{\hat{x}}^{MMSE} & = & \sum_{\supp{S} \in \Omega} \left[\prod_{i\in \supp{S}}g_{i}
\prod_{j\notin \supp{S}}\left(1-g_{j}\right) c^2 \cdot \left( \sum_{k=1}^n
\mats{I}{S}(k)\beta_k\ve{e}_k \right)\right] \nonumber \\
& = & c^2 \sum_{k=1}^n  \left[\sum_{\supp{S} \in \Omega} \mats{I}{S}(k)
\prod_{i\in \supp{S}}g_{i} \prod_{j\notin \supp{S}}\left(1-g_{j}\right) \right]
\beta_k \ve{e}_k.
\end{eqnarray}
We introduce now another observation, similar to the one posed in
Proposition \ref{prop1}. This will be used to further simplify the
above expression.
\begin{proposition}
Let $\Omega$ be the set of all possible subsets of $n$ indices,
and let $g_i$ be values associated with each index, such that $0
\leq g_i \leq 1$. Then,
\begin{equation}
\sum_{\supp{S} \in \Omega} \mats{I}{S}(k) \prod_{i \in \supp{S}} g_i
\cdot \prod_{j \notin \supp{S}} \left(1 - g_j\right) = g_k.
\end{equation}
\label{prop2}
\end{proposition}

\begin{pf}
In the spirit of the coin tossing
interpretation described in the proof of Proposition \ref{prop1},
the multiplication by the expression $\mats{I}{S}(k)$ implies that
only toss outcomes where the $k^{th}$ coin turns up ``heads'' are
included in the summation. Thus, the overall probability of those is
exactly the probability that the $k^{th}$ coin turn up ``heads'',
which is $g_k$ as claimed. A somewhat more formal way to pose this
rationale is by observing that
\begin{eqnarray}
\nonumber \sum_{\supp{S} \in \Omega} \mats{I}{S}(k) \prod_{i \in
\supp{S}} g_i \cdot \prod_{j \notin \supp{S}} \left(1 - g_j\right) &
= & \sum_{\supp{S} \in \Omega~s.t.~k\in \supp{S}} ~~~\prod_{i \in
\supp{S}} g_i \cdot \prod_{j \notin \supp{S}} \left(1 - g_j\right)
\\ \nonumber & = & g_k \cdot \sum_{\supp{S} \in \Omega_k
} ~~~\prod_{i \in \supp{S}} g_i \cdot \prod_{j \notin \supp{S}}
\left(1 - g_j\right).
\end{eqnarray}
The last summation is over the set $\Omega_k$, that contains all the supports in $\Omega$
and do not contain the $k^{th}$ entry. Thus, for the remaining $n-1$
elements, this summation is complete, just as posed in Proposition
\ref{prop1}, and therefore the overall expression equals $g_k$.
\hfill $\Box$
\end{pf}

\begin{figure}
\centering \subfloat[$\sigma =
0.1$]{\includegraphics[width=2.5in]{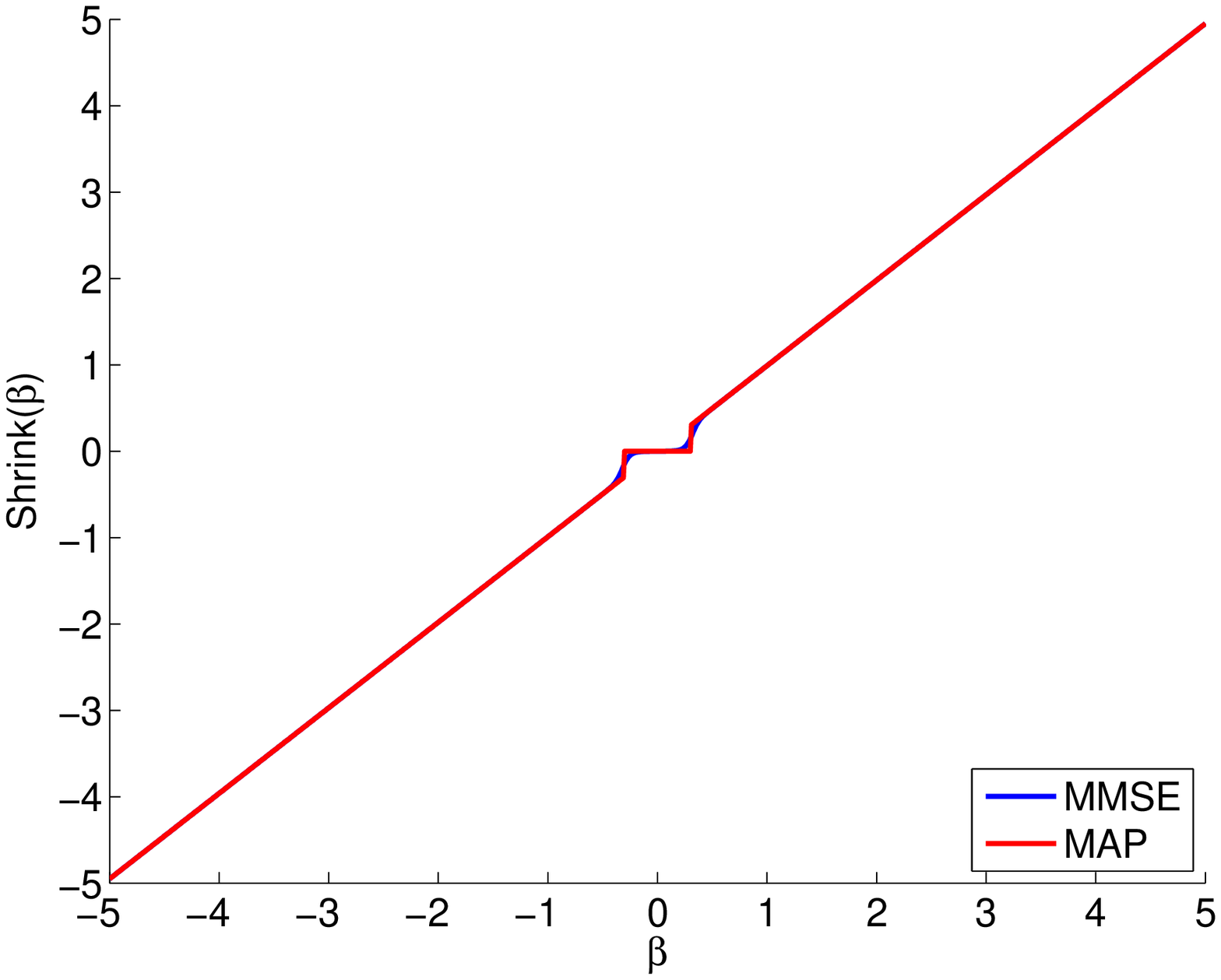}} \qquad
\subfloat[$\sigma = 0.5$]{\includegraphics[width=2.5in]{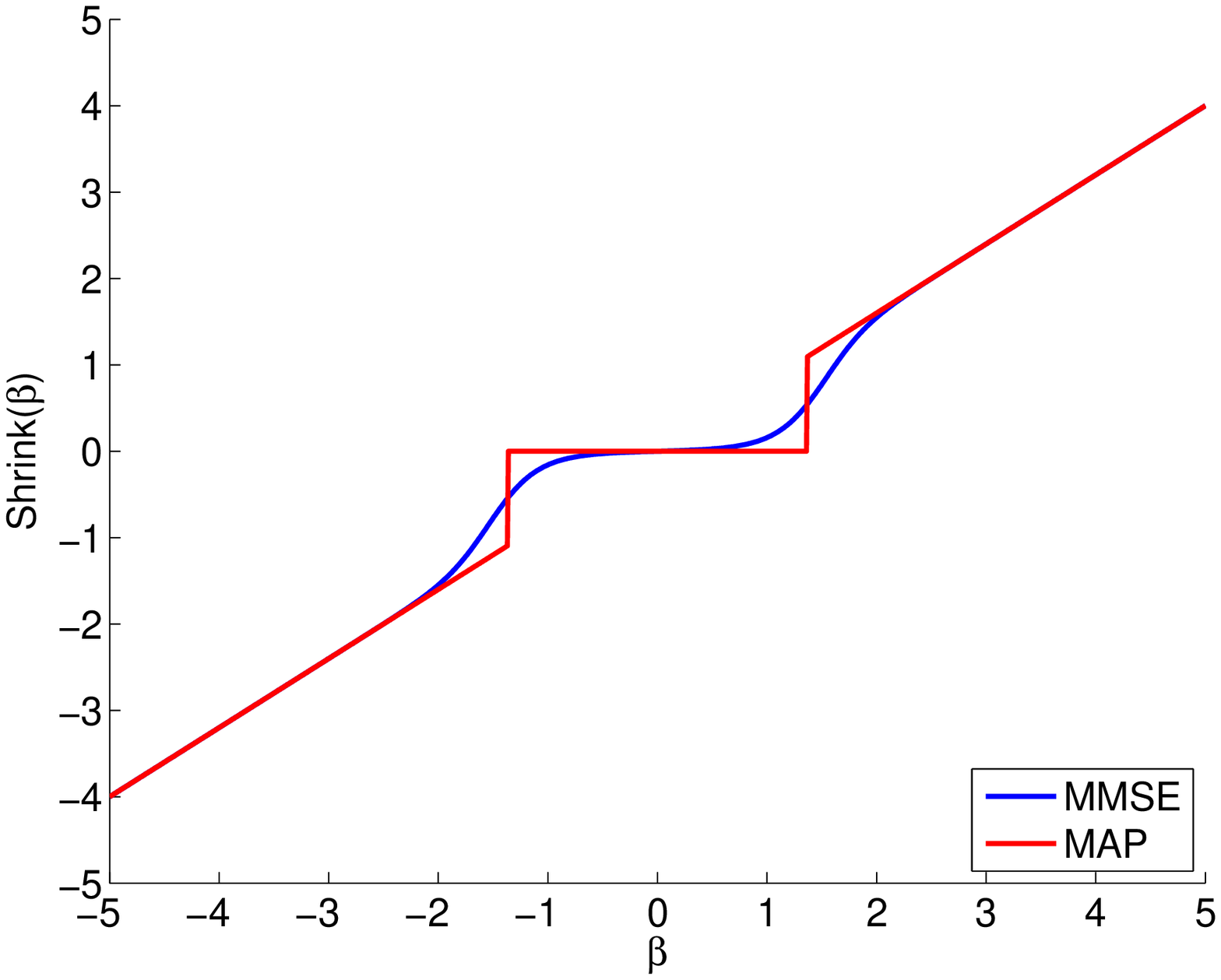}}\\
\subfloat[$\sigma =
1$]{\includegraphics[width=2.5in]{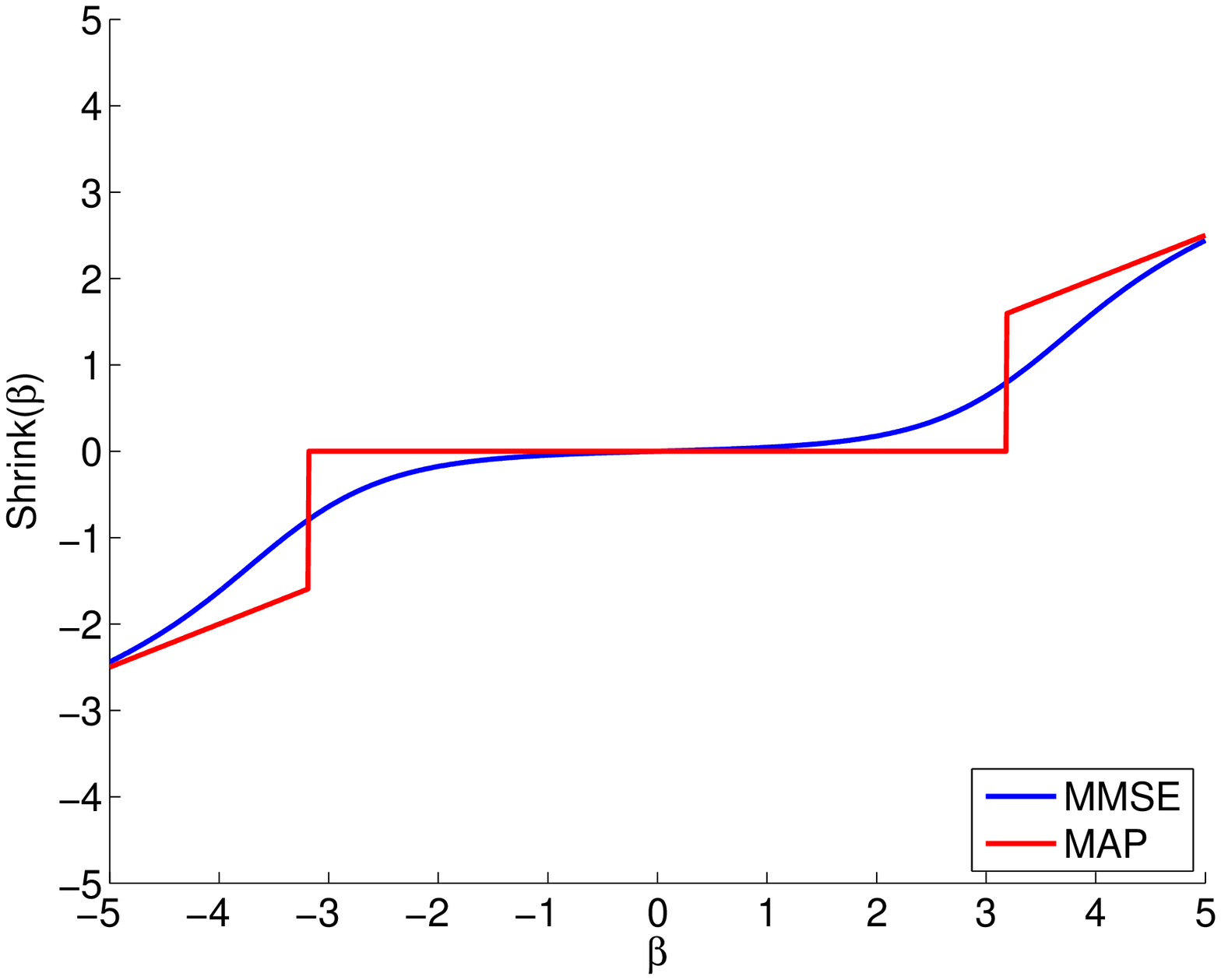}}
\caption{Shrinkage functions for MMSE and MAP estimators ($P_i=0.1$,
$\sigma_x=1$).} \label{fig:shrinkage}
\end{figure}

Returning to the MMSE expression in Equation
\eqref{eq:xMMSE-Unitary}, and using this equality, we get a
far simpler MMSE expression of the form
\begin{eqnarray}\label{eq:MMSE-Unitary-Closeform}
\ve{\hat{x}}^{MMSE} = c^2 \sum_{k=1}^n  g_k \beta_k \ve{e}_k = c^2 \sum_{k=1}^n
\frac{q_k}{1+q_k} \beta_k \ve{e}_k.
\end{eqnarray}
This is an explicit formula for MMSE estimation. The estimation is
computed by first calculating $\ve{\beta}=\mat{D}^T\ve{y}$, and
then simply multiplying each entry $\beta_k$ by $c^2 q_k/(1+q_k)$
(which is a function of $\beta_k$ as well). Explicitly, the MMSE
estimate is given elementwise by
\begin{eqnarray}\label{eq:MMSEshrinkage}
{\hat x}_i^{MMSE} = \psi_{MMSE}(\beta_i) =
\frac{\exp\left\{\frac{c^2}{2\sigma^2}\ve{\beta}_{i}^2\right\}
\frac{P_i}{1-P_i}\sqrt{1-c^2}}{1+\exp\left\{\frac{c^2}{2\sigma^2}\ve{\beta}_{i}^2\right\}
\frac{P_i}{1-P_i}\sqrt{1-c^2}}\cdot c^2 \beta_i.
\end{eqnarray}
This operation has the form of a scalar shrinkage operation, just
like MAP. For $|\beta_i| \ll \sigma/c$ this formula leads to
${\hat x}_i^{MMSE} \approx 0$, whereas for $|\beta_i| \gg
\sigma/c$ the outcome is ${\hat x}_i^{MMSE} \approx c^2\beta_i$
(just like the MAP). Thus, the expression multiplying $c^2\beta_i$
here serves as a soft-shrinkage\footnote{This should not be
confused with the term soft-thresholding obtained when minimizing
an $\ell_1$ penalty.} operation, which replaces the hard-shrinkage
practiced in the MAP. Figure \ref{fig:shrinkage} shows the various
shrinkage functions obtained for each estimator.

\section{Performance Analysis}
\label{sec:Bounds}

\subsection{Deriving the Estimators' MSE}
Our main goal in this work is to develop error expressions for the
different estimators in the unitary regime, exploiting the general
derivations of section \ref{sec:Performance}. We start by
calculating the error for an oracle solution
$\ves{\hat{x}}{S}^{Oracle}$. Using Equation
\eqref{eq:matrixQ-Unitary} we obtain
\begin{equation}
\expec{\norm{\ves{\hat{x}}{S}^{Oracle} - \ve{x}}} = \trace{\mats{Q}{S}^{-1}} =
\suppsize{S}c^2\sigma^2 = \sum_{k=1}^{n}\mats{I}{S}(k)c^2\sigma^2,
\label{eq:traceInvQ}
\end{equation}
where the indicator function is the same as previously used in
\eqref{eq:betaS}. The last equality will become useful for our
later development.

Turning to the MMSE estimator, recall the general expected-MSE
expression in Equation \eqref{eq:general-MSE},
\begin{equation}
\expec{\norm{\ve{\hat{x}}^{MMSE} - \ve{x}}} =
\sum_{\supp{S}\in\Omega} P(\supp{S}|\ve{y}) \cdot \left[
\suppsize{S}c^2\sigma^2 + \norm{\ve{\hat{x}}^{MMSE} -
\ves{\hat{x}}{S}^{Oracle}} \right]. \label{eq:MSE-MMSE-Unit-1}
\end{equation}
Using the unitary MMSE estimator expression in Equation
\eqref{eq:MMSE-Unitary-Closeform} and that of the oracle solution in
Equation \eqref{eq:Xoracle-Unitary}, we further develop the second
term in the expression above, and obtain
\begin{eqnarray}
\norm{\ve{\hat{x}}^{MMSE} - \ves{\hat{x}}{S}^{Oracle}} & = & \norm{c^2 \sum_{k=1}^{n}
g_k \beta_k \ve{e}_k - c^2\sum_{k=1}^{n} \mats{I}{S}(k)\beta_k\ve{e}_k} \nonumber \\
& = & \sum_{k=1}^{n}  c^4 \left[ g_k - \mats{I}{S}(k) \right]^2 \beta_k^2 \nonumber \\
& = & \sum_{k=1}^{n}  c^4 \left[ g_k^2 - 2g_k\mats{I}{S}(k) +
\mats{I}{S}(k) \right] \beta_k^2. \label{eq:normMMSE-OR}
\end{eqnarray}
Plugging this expression back into Equation
\eqref{eq:MSE-MMSE-Unit-1}, together with the expression for
$P(\supp{S}|\ve{y})$ in Equation \eqref{eq:SgivenY-Unitary-4}, gives
\begin{eqnarray}
\MSE{\ve{\hat{x}}^{MMSE}} & = & \sum_{\supp{S}\in\Omega} P(\supp{S}|\ve{y})
\left\{c^2\sigma^2\sum_{k=1}^{n} \mats{I}{S}(k) +  c^4 \sum_{k=1}^{n}
\left[ g_k^2 - 2g_k\mats{I}{S}(k) + \mats{I}{S}(k) \right] \beta_k^2 \right\} \nonumber \\
& = & \sum_{k=1}^{n} \left\{ c^2\sigma^2 \sum_{\supp{S}\in\Omega}
\mats{I}{S}(k)\prod_{i\in \supp{S}}g_{i}\prod_{j\notin
\supp{S}}\left(1-g_{j}\right) \right. \nonumber \\
& & \left. + c^4 \beta_k^2 \sum_{\supp{S}\in\Omega} ~~\prod_{i\in
\supp{S}}g_{i}\prod_{j\notin \supp{S}}\left(1-g_{j}\right)
\left[ g_k^2 - 2g_k\mats{I}{S}(k) + \mats{I}{S}(k) \right]  \right\} \nonumber \\
& = & \sum_{k=1}^{n} c^2\sigma^2 g_k + c^4 \beta_k^2 \left(g_k -
g_k^2 \right). \label{eq:MMSE-error-unitary}
\end{eqnarray}
Here we have exploited Proposition \ref{prop2}. Interestingly, the
property $|\supp{S}|=\sum_{k=1}^{n} \mats{I}{S}(k)$ and
Proposition \ref{prop2} yield the relationship
\begin{eqnarray}
\expec{\suppsize{S}} & = & \sum_{\supp{S}\in\Omega} P(\supp{S}|y) \suppsize{S}  \nonumber \\
& = & \sum_{\supp{S}\in\Omega} \left(\sum_{k=1}^{n} \mats{I}{S}(k)
\right)\prod_{i\in \supp{S}}g_{i}\prod_{j\notin \supp{S}}
\left(1-g_{j}\right) \nonumber \\
& = & \sum_{k=1}^{n} \left[ \sum_{\supp{S}\in\Omega}  \mats{I}{S}(k)
\prod_{i\in \supp{S}}g_{i}\prod_{j\notin \supp{S}}
\left(1-g_{j}\right) \right] = \sum_{k=1}^{n} g_k.
\end{eqnarray}
This implies that the MMSE error can be alternatively written as
\begin{eqnarray}
\MSE{\ve{\hat{x}}^{MMSE}} =c^2\sigma^2 \expec{\suppsize{S}} + c^4
\sum_{k=1}^{n}  \beta_k^2 \left(g_k - g_k^2 \right),
\label{eq:MMSE-error-unitary2}
\end{eqnarray}
suggesting that the error is composed of an ``oracle''
error\footnote{See the similarity between the first term here and
the one posed in Equation \eqref{eq:traceInvQ}.}, and an
additional part that is necessarily positive (since $0<g_k<1$). As
an extreme example, if the elements of the vector $\ve{\beta}$
tend to be either very high or very low (compared to $\sigma/c$),
then the $g_k$ tend to the extremes as well. In such a case, the
second term nearly vanishes, and the performance is close to that
of the oracle.

We next study the MAP performance. Recall Equation
\eqref{eq:general-MSE}, and note that $\ve{\hat{x}}^{MAP}$ may be
written as
\begin{equation}
\ve{\hat{x}}^{MAP} = \sum_{k=1}^{n} \mats{I}{MAP}(k) c^2 \ve{\beta}_{k} \ve{e}_{k},
\label{eq:xMAP-indicator}
\end{equation}
where $\mats{I}{MAP}(k)$ is an indicator function for the MAP
support. Exploiting  Propositions \ref{prop1} and \ref{prop2}, we
obtain the following expression for the MAP mean-squared-error,
\begin{eqnarray}
\MSE{\ve{\hat{x}}^{MAP}} & = & \sum_{\supp{S}\in\Omega}
P(\supp{S}|\ve{y}) \left\{c^2\sigma^2\sum_{k=1}^{n} \mats{I}{S}(k) +
c^4 \sum_{k=1}^{n}
\left[ \mats{I}{MAP}(k) - 2\mats{I}{S}(k)\mats{I}{MAP}(k) + \mats{I}{S}(k) \right] \beta_k^2 \right\} \nonumber \\
& = & \sum_{k=1}^{n} \left\{ c^2\sigma^2 \sum_{\supp{S}\in\Omega}
\mats{I}{S}(k)\prod_{i\in \supp{S}}g_{i}\prod_{j\notin
\supp{S}}\left(1-g_{j}\right) \right. \nonumber \\
& & \left. + c^4 \beta_k^2 \sum_{\supp{S}\in\Omega} ~~\prod_{i\in
\supp{S}}g_{i}\prod_{j\notin \supp{S}}\left(1-g_{j}\right)
\left[ \mats{I}{MAP}(k) + \mats{I}{S}(k) - 2\mats{I}{S}(k)\mats{I}{MAP}(k) \right]  \right\} \nonumber \\
& = & \sum_{k=1}^{n} c^2\sigma^2 g_k + c^4 \beta_k^2 \left[g_k + \mats{I}{MAP}(k)(1 - 2g_k) \right].
\label{eq:MAP-error-unitary}
\end{eqnarray}
Analyzing the difference between the MMSE and MAP errors, in
Equations \eqref{eq:MMSE-error-unitary} and
\eqref{eq:MAP-error-unitary} respectively, we find that only the
last terms in each are different: $-g_k^2$ versus
$\mats{I}{MAP}(k)(1-2g_k)$, respectively. Obviously, this implies
$\MSE{\ve{\hat{x}}^{MMSE}} \le \MSE{\ve{\hat{x}}^{MAP}}$, because
$-g_k^2 \le \mats{I}{MAP}(k)(1-2g_k)$ for any $k$, and regardless of
the value of $\mats{I}{MAP}(k)$ (zero or one).

In order to further understand the estimators'  performance given in
the Equations \eqref{eq:MMSE-error-unitary} and
\eqref{eq:MAP-error-unitary}, we turn  now to a further analysis of
these expressions and derive worst-case upper-bounds for them.
The bounds we are about to build do not depend on the dimension of
the signal, but rather on the problem parameters ($\sigma, \sigma_x,
P_i$) alone. We begin with the MMSE, then turn to the MAP, and
finally compare and discuss the resulting bounds.


\subsection{MMSE Performance Bound}

Referring to Equation \eqref{eq:MMSE-error-unitary}, which describes
the error associated with the MMSE approximation, we shall denote by
$MSE_1$ the first term,
\begin{equation}
MSE_1 = c^2 \sigma^2 \sum_{k=1}^n g_k. \label{eq:MSE1}
\end{equation}
As mentioned before, this is the expected MSE of the oracle (given
\ve{y}). The second term, denoted by $MSE_2$, is given by
\begin{equation}
MSE_2 = c^4 \sum_{k=1}^n \beta_k^2 g_k (1-g_k).
\end{equation}
This is the additional error due to the fact that the support is
unknown. We would like to bound the ratio $r = MSE_2/ MSE_1$, as
this immediately yields a bound ($r+1$) on the MMSE error in terms
of the expected oracle error. Our goal is thus to characterize the
worst ratio
\begin{equation}
\max_{\ve{y} \in R^n} r = \max_{\ve{y} \in R^n} \frac{MSE_2}{MSE_1},
\label{eq:rm-ratio}
\end{equation}
that is, the worst (largest) ratio over all conceivable signals
\ve{y}, where the dependence on \ve{y} enters via the
$\ve{\beta}_k$'s. In order to characterize this ratio, we shall need
the following simple lemma:


\begin{lemma} \label{lem:fractions}
Let $\left(a_k,b_k\right), k=1,\dots,n$, be pairs of positive real
numbers. Let $m$ be the index of a pair whose ratio is maximal,
i.e.,
\begin{equation} \protect\label{eq:Lemma_Inequalities}
\frac{a_k}{b_k} \leq \frac{a_m}{b_m} ~~~~~\mbox{for all} ~k \geq 1.
\end{equation}
Then
\begin{equation*}
\frac{\sum_{k=1}^n a_k}{\sum_{j=1}^n b_j} \leq \frac{a_m}{b_m},
\end{equation*}
with equality occurring only if $\frac{a_1}{b_1} = \frac{a_2}{b_2} =
\dots = \frac{a_n}{b_n}$.
\end{lemma}

\begin{pf}
By (\ref{eq:Lemma_Inequalities}), $a_k b_m
\leq a_m b_k$ for all $k \geq 1$, with equality obtained only if
$a_k/b_k = a_m/b_m$. Summing up all these inequalities, we obtain
\begin{equation*}
b_m \sum_{k=1}^n a_k  \leq a_m \sum_{j=1}^n b_j \, ,
\end{equation*}
hence,
\begin{equation*}
 \frac{\sum_{k=1}^n a_k}{\sum_{j=1}^n b_j}  \leq \frac{a_m}{b_m},
\end{equation*}
as claimed, with equality occurring only if $\frac{a_i}{b_i} =
\frac{a_m}{b_m}$, for every $i$. \hfill $\Box$
\end{pf}

Returning to our task of bounding $MSE_2/MSE_1$, we observe that
this ratio can be written as
\begin{equation}\label{eq:MSE2-1}
\frac{MSE_2}{MSE_1} = \frac{c^4 \sum_{k=1}^n \beta_k^2 g_k
(1-g_k)}{c^2 \sigma^2 \sum_{k=1}^n g_k},
\end{equation}
which is of the same form as the ratio appearing in the Lemma. This leads us
to the following Theorem:


\begin{theorem} \label{th:MMSE-Bound}
Denote $G_k = \sqrt{1-c^2}P_k/(1-P_k)$, and let $m$ be the index
corresponding to an a priori least likely atom, i.e., $P_m =
\min_{1 \leq k \leq n} P_k$ and hence, $G_m = \min_{1 \leq k \leq
n} G_k$. Denote $f_{MMSE}(s) = \frac{2s}{1+G_m e^s}$, and define
(implicitly) $s^{\star} = \arg\max_{s\geq 0} f_{MMSE}(s)$. Then
\begin{enumerate}
\item $r^{\star} = f_{MMSE}(s^{\star})$ is an upper-bound on the
ratio $MSE_2 / MSE_1$.

\item The worst ratio, $r^{\star}$, satisfies the explicit bound
\begin{eqnarray}\label{eq:r_bar}
r^{\star} \le \left\{ \begin{array}{cc} 2\ln \left(
\frac{1}{4G_m}\right) & G_m < \frac{1}{4e^2}\approx 0.034 \\
\frac{2}{\sqrt{G_m} e} & G_m \ge \frac{1}{4e^2}\approx 0.034
\end{array}. \right.
\end{eqnarray}
\end{enumerate}
\end{theorem}


\begin{pf}
Starting with the first claim, we embark from
Equation \eqref{eq:MSE2-1} and exploit Lemma \ref{lem:fractions} to
obtain
\begin{eqnarray}\label{eq:thm_step1}
\frac{MSE_2}{MSE_1} & = & \frac{c^4 \sum_{k=1}^n \beta_k^2 g_k
(1-g_k)}{c^2 \sigma^2 \sum_{k=1}^n g_k} \nonumber \\  & \le &
\frac{c^2}{\sigma^2}\cdot \max_{1\le k\le n} \frac{\beta_k^2 g_k
(1-g_k)}{g_k} \nonumber \\  & \le & \frac{c^2}{\sigma^2}\cdot
\max_{1\le k\le n} \beta_k^2 (1-g_k).
\end{eqnarray}
Recalling that $g_k=q_k/(1+q_k)$, the definition of $q_k$ in
\eqref{eq:def_qi}, and the definition of $G_k$ above, we have
\begin{eqnarray}\label{eq:thm_step2}
1-g_k =\frac{1}{1+q_k} & = & \frac{1}{1+
\frac{P_k}{1-P_k}\sqrt{1-c^2}
\exp\left\{\frac{c^2}{2\sigma^2}\ve{\beta}_{k}^2\right\} } \nonumber
\\ &  = & \frac{1}{1+G_k \exp\left\{c^2
\beta_k^2/2\sigma^2\right\}}.
\end{eqnarray}
Plugging this into Equation \eqref{eq:thm_step1} and denoting
$s=c^2 \beta_k^2/2\sigma^2$, we obtain
\begin{eqnarray}\label{eq:thm_step3}
\frac{MSE_2}{MSE_1} \le  \max_{1\le k\le n} \frac{2s}{1+G_k \exp\{
s \}}.
\end{eqnarray}
This is a monotonically decreasing function of $G_k$ for any fixed
value of $s \ge 0$ (note that $s$ must be non-negative, due to its
definition). Thus, the maximum over the indices $1\le k \le n$ is
obtained for the index $m$ for which $G_k$ is the smallest.
Therefore,
\begin{eqnarray}\label{eq:thm_step4}
\max_{\ve{\beta}}\frac{MSE_2}{MSE_1} \le \max_{s\ge
0}\frac{2s}{1+G_m \exp \{s \}} = f_{MMSE}(s^\star) = r^\star,
\end{eqnarray}
as claimed.

Turning to the second claim of the theorem, we desire to bound
$f_{MMSE}(s)$ from above. To this end, we maximize the alternative
function $\overline{f}(s)$ that bounds $f_{MMSE}(s)$ from above
point-wise:
\begin{equation} \label{eq:Upper_Bound_Function}
f_{MMSE}(s) = \frac{2s}{1+G_m e^s} \leq \frac{2s}{\max\left( 1,
2\sqrt{G_m e^s }\right)} \equiv \overline{f}(s).
\end{equation}
Here we have used the facts that (i) the arithmetic mean $( 1+G_m
e^s)/2 $ is necessarily larger than the geometric one, $\sqrt{G_m
e^s }$, and (ii) $1+G_m e^s \ge 1$.

The switch-over in the denominator of $\overline{f}(s)$ occurs
when $G_me^s = 1/4$, which takes place for $s=s_0 \equiv  \ln
(1/4G_m)$. For $s \leq s_0$, $\overline{f}(s) = 2s$, which is
monotonically increasing. For $s \geq s_0$, $\overline{f}(s) =
s/\sqrt{G_m e^s}$, whose derivative is given by $\overline{f'}(s)
= (1-s/2)/\sqrt{G_m e^s}$. Thus, if $s_0 \geq 2$, the maximum of
$\overline{f}(s)$ occurs at $s= s_0$, being
$\overline{f}(s_0)=2s_0=2\ln (1/4G_m)$. Otherwise, the maximum
occurs at $s=2$, being $\overline{f}(2)=2/\sqrt{G_m} e$. This
proves the explicit upper bound on $r^{\star}$, as given in
Equation \eqref{eq:r_bar}. \hfill $\Box$
\end{pf}

\begin{figure}
   \centering
   \includegraphics[width=4in]{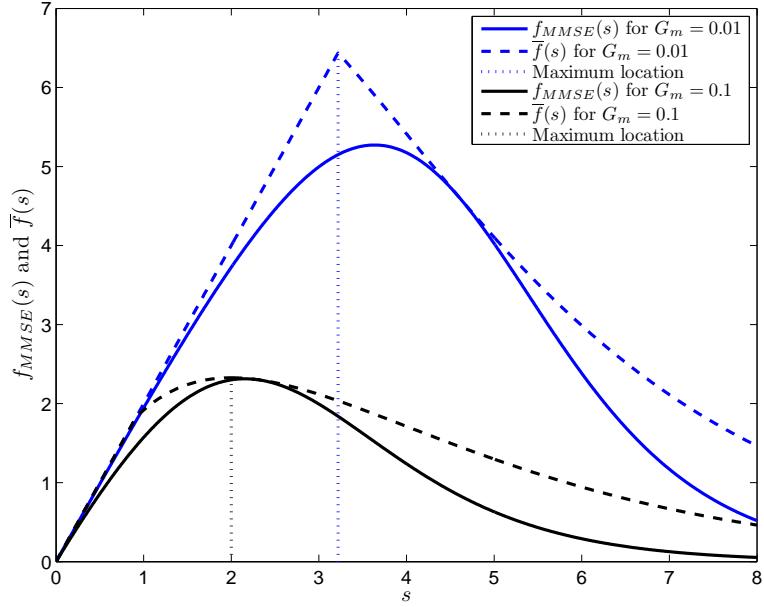}
   \caption{Graph plot of the function $f_{MMSE}(s)$ and its
   upper-bounding function $\overline{f}(s)$ (the solid and the dashed lines,
   respectively), exhibiting the two cases, where the maximum
   changes given the value $G_m$.}
   \label{fig:f-mmse1}
\end{figure}

\begin{figure}
   \centering
   \includegraphics[width=4in]{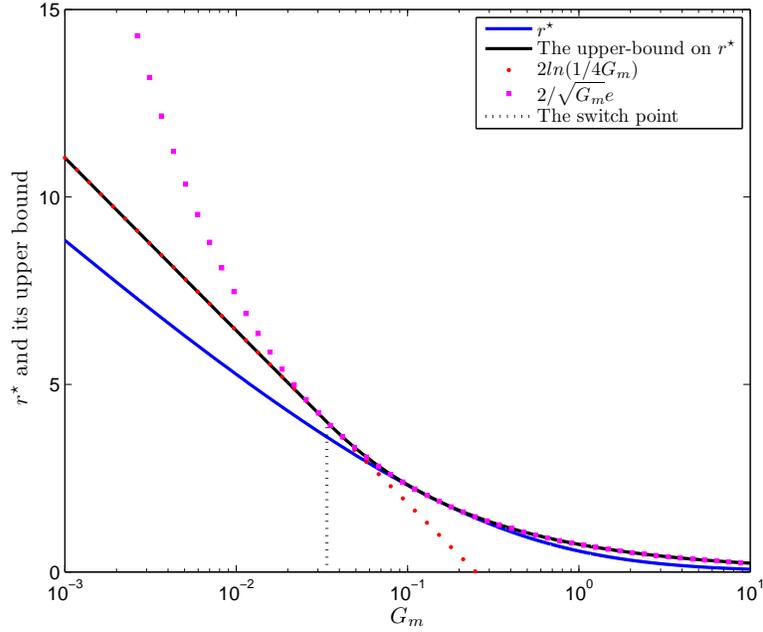}
   \caption{The worst ratio $r^\star$ and its upper bound, as
   given in Equation \eqref{eq:r_bar}. This graph also shows the two
   portions of this bounding function, and the location of the
   switch between them.}
   \label{fig:f-mmse2}
\end{figure}

Figure \ref{fig:f-mmse1} shows the functions $f_{MMSE}(s)$ and its
upper bound $\overline{f}(s)$ for two possible values of $G_m$:
$0.01$ and $0.1$. These two cases correspond to the two options
covered in Equation \eqref{eq:r_bar}. As can be seen, for
$G_m=0.01 <0.034$, the maximum point is obtained on the linear
part of $\overline{f}(s)$, whereas in the case of $G_m=0.1
>0.034$, the maximum is obtained for $s=2$. Figure
\ref{fig:f-mmse2} presents the value of $r^\star$ as a function of
$G_m$. This figure also shows the upper-bound on this value as
given in Equation \eqref{eq:r_bar}, and the two sub-functions that
comprise it.

\begin{corollary}
The expected error for the MMSE estimator is bounded for any signal
\ve{y} by
\begin{equation}
\MSE{\ve{\hat{x}}^{MMSE}} \leq \MSE{\ve{\hat{x}}^{Oracle}}\cdot
\left\{
\begin{array}{lc}
 1 + 2\ln\frac{1}{4G_m} &  G_m \leq \frac{1}{4} e^{-2}\\
 1 + \frac{2}{\sqrt{G_m}e} &  G_m \geq \frac{1}{4} e^{-2}
 \end{array}
\right. .
\end{equation}
\end{corollary}

\begin{pf}
Follows from Theorem \ref{th:MMSE-Bound}.
\hfill $\Box$
\end{pf}

What happens when all the probabilities $P_k$ are equal? In such a
case we obtain that $G_1=G_2= ~\cdots ~=G_n$. From Equation
\eqref{eq:thm_step1}, which uses Lemma \ref{lem:fractions}, it is
obvious that the worst-ratio $r^\star$ becomes a tight upper-bound
on $MSE_2/MSE_1$, since all the terms in the numerator and the
denominator summations are equal. Furthermore, the worst-case
$\beta_k$'s are all equal to $\pm \sigma\sqrt{s^\star}/c$.


\subsection{MAP Performance Bound}

We next develop an upper-bound on the error associated with the
MAP estimate in Equation \eqref{eq:MAP-error-unitary}. While
$MSE_1$ remains the same as in Equation \eqref{eq:MSE1}, the term
that corresponds to $MSE_2$ for the MAP becomes
\begin{equation}
MSE_2 = c^4 \sum_{k=1}^n \beta_k^2 g_k \left[ 1 +
\frac{\mats{I}{MAP}(k)(1-2g_k)}{g_k} \right]. \nonumber
\end{equation}
Continuing with the same definitions  as in the previous section,
we prove a similar theorem for the expected MSE  of the MAP
estimator.

\begin{theorem} \label{th:MAP-Bound}
Denote $G_k = \sqrt{1-c^2}P_k/(1-P_k)$, and let $m$ be the index
corresponding to an a priori least likely atom, i.e., $P_m =
\min_{1 \leq k \leq n} P_k$ and hence, $G_m = \min_{1 \leq k \leq
n} G_k$. Define the function
\begin{eqnarray}
f_{MAP}(s) = \left\{
\begin{array}{lr} 2s  & G_me^s<1 \\
\frac{2s}{G_m e^s} & G_me^s \geq 1
\end{array} \right.,
\end{eqnarray}
and define (implicitly) $s^\star = \arg\max_{s\geq 0} f_{MAP}(s)$.
Then
\begin{enumerate}
\item $r^{\star} = f_{MAP}(s^{\star})$ is an upper-bound on the
ratio $MSE_2 / MSE_1$. \item The worst ratio, $r^{\star}$,
satisfies the explicit bound
\begin{eqnarray}\label{eq:r_bar_map}
r^{\star} = \left\{
\begin{array}{cc}
    2\ln{\frac{1}{G_m}} & G_m < e^{-1}    \approx 0.368 \\
    \frac{2}{G_me}      & G_m \geq e^{-1} \approx 0.368
\end{array}. \right.
\end{eqnarray}
\end{enumerate}
\end{theorem}

\begin{pf}
The proof follows the same lines as that of
Theorem \ref{th:MMSE-Bound}. Starting with the ratio $r$, we
exploit Lemma \ref{lem:fractions} and obtain
\begin{eqnarray}
\frac{MSE_2}{MSE_1} & = & \frac{c^4\sum_{k=1}^n\beta_k^2 g_k
\left[ 1 + \mats{I}{MAP}(k)\frac{1-2g_k}{g_k} \right]}{c^2
\sigma^2 \sum_{k=1}^n g_k} \nonumber \\
&\leq& \frac{c^2}{\sigma^2}\cdot \max_{1\le k\le n}
\frac{\beta_k^2g_k\left[ 1 + \mats{I}{MAP}(k)\frac{1-2g_k}{g_k}
\right]}{g_k} \nonumber \\ &\leq& \frac{c^2}{\sigma^2}\cdot
\max_{1\le k\le n} \beta_k^2\left[ 1 +
\mats{I}{MAP}(k)\frac{1-2g_k}{g_k} \right]. \label{eq:thm2_step1}
\end{eqnarray}
Again using the relation $g_k=q_k/(1+q_k)$ and the definition of
$q_k$ from \eqref{eq:def_qi}, we have that
\begin{eqnarray}
\frac{1-2g_k}{g_k} = \frac{1}{q_k} - 1 = \frac{1}{G_k
\exp\left\{\frac{c^2\beta_k^2}{2\sigma^2}\right\}}-1 =
\frac{1}{G_k \exp \{s\}}-1, \label{eq:thm2_step2}
\end{eqnarray}
where we have used the definition of $s$ as before
($s=c^2\beta_k^2/2\sigma^2$). Plugged back into Equation
\eqref{eq:thm2_step1}, we obtain
\begin{eqnarray}\label{eq:thm2_step3}
\frac{MSE_2}{MSE_1}  \le  \max_{1\le k\le n} 2s \left[ 1 +
\mats{I}{MAP}(k) \left(\frac{1}{G_k \exp\{s\}}-1 \right) \right].
\end{eqnarray}
For any fixed value of $s$, the maximum over the indices $1\le k
\le n$ is obtained for the index $m$ for which $G_k$ is the
smallest. Therefore, maximizing this expression with respect to
both $k$ and $s$ yields
\begin{eqnarray}\label{eq:thm2_step4}
\max_{\beta}\frac{MSE_2}{MSE_1} & \le & \max_{s\geq 0}
2s\left[ 1 + \mats{I}{MAP}(m) \left(\frac{1}{G_m \exp\{s\}}-1 \right) \right] \\
& \le & \nonumber \max_{s\geq 0} \left\{ \begin{array}{cc}
2s & G_m \exp\{s\} < 1 \\
\frac{2s}{G_me^{s}} & G_m\exp\{s\} \geq 1
\end{array}, \right. .
\end{eqnarray}
Here we have used the fact that $\mats{I}{MAP}(m)=1$ when the atom
$m$ is part of the MAP support, which takes place if $q_m \geq
1$ (see the discussion after Equation \eqref{eq:MAP-Unitary}).

We turn to the second claim of the theorem, and calculate
explicitly the value $s^{\star}$ for which
$f_{MAP}(s^{\star})=r^{\star}$ is maximized. The switch-over
between the two cases of $f_{MAP}(s)$ occurs when $G_me^s = 1$,
that is, $s=s_0 \equiv \ln \left( 1/G_m \right)$. For $s \leq
s_0$, $f_{MAP}(s) = 2s$, which is monotonically increasing. For $s
\geq s_0$, $f_{MAP}(s) = 2s/(G_m e^s)$, whose derivative is given
by $f'(s) = (2-2s)/(G_m e^s)$. Thus, if $s_0 > 1$, the maximum of
$f$ occurs at $s^{\star} = s_0$, that is, $f_{MAP}(s_0)=2\ln
(1/G_m)$. Otherwise, the maximum occurs at $s^{\star}=1$ with
$f_{MAP}(1)=2/(G_me)$. This proves the explicit upper bound
$r^{\star}$ as given in Equation \eqref{eq:r_bar_map}. \hfill
$\Box$
\end{pf}

\begin{figure}
   \centering
   \includegraphics[width=4in]{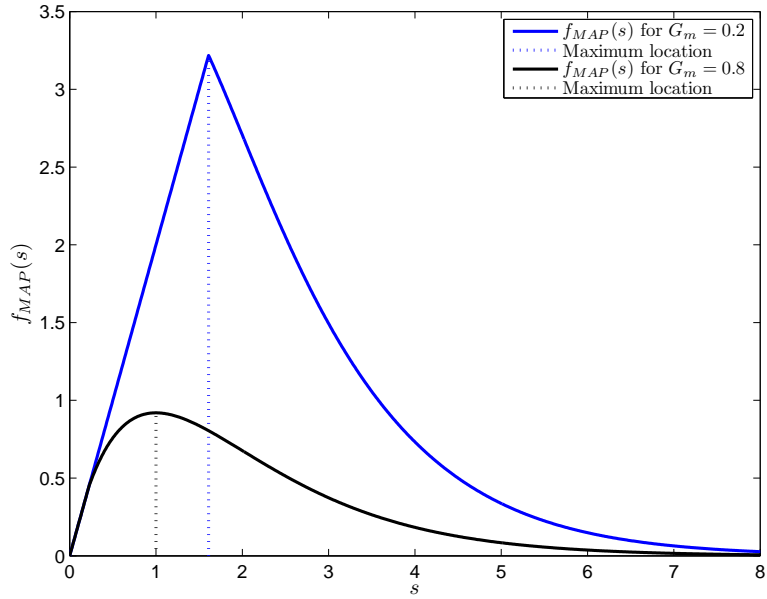}
   \caption{A plot of the function $f_{MAP}(s)$,
   exhibiting the two cases, where the maximum
   changes character according to the value $G_m$.}
   \label{fig:f-map1}
\end{figure}

Figure \ref{fig:f-map1} shows two examples of $f_{MAP}(s)$ for two
possible values of $G_m$: $0.2$ and $0.8$. These two cases
correspond to the two options covered in Equation
\eqref{eq:r_bar_map}. As can be seen, for $G_m=0.2 < 0.368$, the
maximum point of $f_{MAP}$ is obtained at the switch-over point,
whereas in the case of $G_m = 0.8 > 0.368$, the maximum is found
at $s=1$.

\begin{figure}
   \centering
   \includegraphics[width=4in]{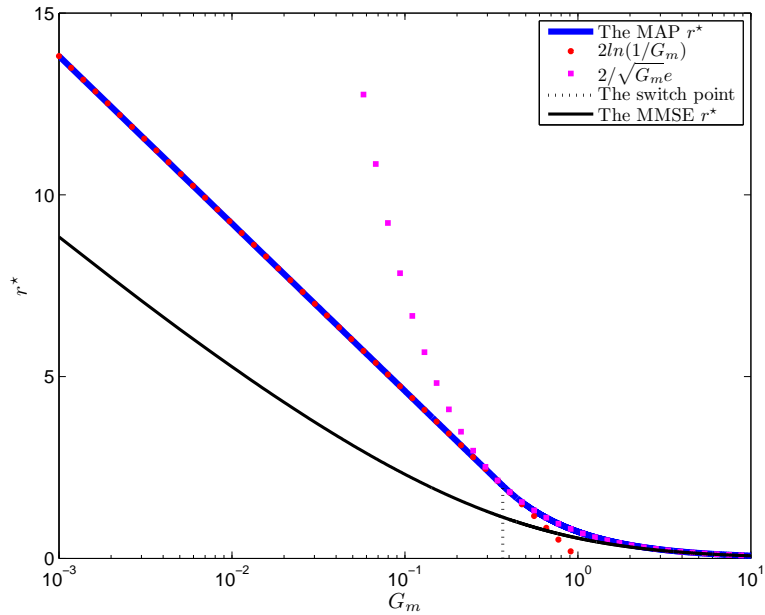}
   \caption{The worst ratio $r^\star$ for MAP, as
   given in Equation \eqref{eq:r_bar_map}. This graph also shows the two
   portions of this function, the location of the
   switch between them, and the MMSE ratio $r^\star$.}
   \label{fig:f-map2}
\end{figure}

Figure \ref{fig:f-map2} presents the value of $r^\star$ as a
function of $G_m$ for both the MAP and the MMSE. This figure also
shows the two sub-functions that construct $r^\star$ for the MAP,
as described in Equation \eqref{eq:r_bar_map}.

\begin{corollary}
The expected MSE error for the MAP estimator is bounded for any
signal \ve{y} by
\begin{equation}
\MSE{\ve{\hat{x}}^{MAP}} \leq \MSE{\ve{\hat{x}}^{Oracle}}\cdot
\left\{
\begin{array}{lc}
 1 + 2\ln\frac{1}{G_m} & G_m \leq e^{-1}\\
 1 + \frac{2}{G_me} & G_m \geq e^{-1}
 \end{array}
\right. .
\end{equation}
\end{corollary}

\begin{pf}
Follows from Theorem \ref{th:MAP-Bound}.
\hfill $\Box$
\end{pf}

When all the probabilities $P_i$ are equivalent, and hence
$G_1=\dots=G_n$, we get again that the worst ratio $r^{\star}$
becomes a tight upper bound on $MSE_2/MSE_1$, following the same
reasoning as explained in the MMSE case. The worst-case
$\beta_k$'s are all given by
\begin{equation*}
\beta_k = \left\{ \begin{array}{lc}
\pm \frac{2\sigma^2}{c^2}\sqrt{2\ln\left(\frac{1}{G_m}\right)} & G_m < e^{-1}\\
\pm \frac{2\sigma^2}{c^2} & G_m \geq e^{-1}
\end{array}. \right.
\end{equation*}

\subsection{MMSE and MAP Bounds -- A Summary}

The bounds developed above suggest that both the MMSE and the MAP
estimators lead in the unitary case to a mean-squared error that
is at worst a constant times the oracle MSE. The analysis given
above provides exact expressions for these ratios.

We should note that the bounds developed above are based on a
worst-case scenario. A more  practical goal would be to bound the
average case, as this should tell us more about the behavior of
real-life signals. We leave this topic to future work.

As a last point in this section, we consider the following
question: When are the MAP and MMSE nearly equivalent? Recall that
the errors of these two estimators are given in Equations
\eqref{eq:MMSE-error-unitary2} and \eqref{eq:MAP-error-unitary} as
\begin{eqnarray*}
\MSE{\ve{\hat{x}}^{MMSE}} & = & \sum_{k=1}^{n} c^2\sigma^2 g_k +
c^4 \sum_{k=1}^{n}  \beta_k^2 \left(g_k - g_k^2 \right) \\
\MSE{\ve{\hat{x}}^{MAP}} & = & \sum_{k=1}^{n} c^2\sigma^2 g_k + c^4
\sum_{k=1}^{n}\beta_k^2 \left[g_k + \mats{I}{MAP}(k)(1 - 2g_k)
\right].
\end{eqnarray*}
In order for these two errors to be close, we should therefore
impose for all $k$
\begin{eqnarray}\label{eq:requirement1}
g_k - g_k^2   \approx g_k + \mats{I}{MAP}(k)(1 - 2g_k)
~~~\Rightarrow~~~g_k^2 - 2\mats{I}{MAP}(k)g_k+\mats{I}{MAP}(k)
\approx 0.
\end{eqnarray}

If $P_k \rightarrow 0$, this leads to $g_k \rightarrow 0$, since
$g_k=q_k/(1+q_k)$ and $q_k=\sqrt{1-c^2}P_k/(1-P_k)\cdot
\exp{c^2\beta_k^2/2\sigma^2}$. From Equation \eqref{eq:MAPshrinkage}
we also have that $\mats{I}{MAP}(k)=0$, implying that this index is
not part of the MAP support. Returning to the requirement posed in
Equation \eqref{eq:requirement1}, we obtain the condition $g_k^2
\approx 0$, which is readily satisfied. Thus, we conclude that one
case where the two estimators, MAP and MMSE, align, is when $P_k
\rightarrow 0$.

When $P_k \rightarrow 1$, this leads to $g_k \rightarrow 1$. Relying
again on Equation \eqref{eq:MAPshrinkage} we also have that
$\mats{I}{MAP}(k)=1$ this time, implying that this index is now part
of the MAP support. Returning to the requirement posed in Equation
\eqref{eq:requirement1}, we obtain the condition $g_k^2 - 2g_k +1 =
(g_k-1)^2\approx 0$, again satisfied (since $g_k$ is close to $1$.
Thus, another case where the two estimators align is when $P_k
\rightarrow 1$.


\section{Experimental Results}
\label{sec:Experiments}

Here we demonstrate the MAP and MMSE estimators for unitary
dictionaries and provide both synthetic and real-signal
experiments to illustrate these algorithms.

\subsection{Synthetic Experiments}

In the first experiment we use a 2D Wavelet dictionary \mat{D}
(Daubachies-$5$ filters) \cite{DAU}, with 3 levels of resolution.
We choose all the atom probabilities $P_i$ and all the variances
$\sigma_i$ to be the same in this test. We use $P=0.1$ and
$\sigma_x=1$.

Generating a two-dimensional signal according to the proposed model
is done by first randomly choosing whether each atom is part of the
support or not with probability $P$. For the selected atoms,
coefficients $x_i$ are drawn independently from a normal
distribution $\normdist(0,\sigma_x^2)$. The resulting sparse vector
of coefficients is multiplied by the unitary dictionary to obtain
the {\em ground-truth} two-dimensional signal.  Each entry is
independently contaminated by white Gaussian noise
$\normdist(0,\sigma^2)$ to create the {\em input} signal \ve{y}. The
values of the additive noise power, $\sigma$, are varied in the
range $[0.1,1]$ to demonstrate the effect of the noise level on the
overall performance. Each of the (noisy) signals is then
approximated using the following estimators:
\begin{enumerate}
\item \textbf{Empirical Oracle} estimation and its MSE. This
estimator appears in Equation \eqref{eq:Xoracle-Unitary}.

\item \textbf{Theoretical Oracle} estimation error, as given in Equation
\eqref{eq:traceInvQ}.

\item \textbf{Empirical MMSE} estimation and
its MSE. We use Equation \eqref{eq:MMSE-Unitary-Closeform} in order
to compute the estimation, and then assess its error empirically.

\item \textbf{Theoretical MMSE}
estimation error, using Equation \eqref{eq:MMSE-error-unitary}
directly.

\item \textbf{Empirical MAP} estimation and its MSE. We
use the closed-form solution given in Equation
\eqref{eq:MAPshrinkage}.

\item \textbf{Theoretical MAP} estimation error, as given in Equation
\eqref{eq:MAP-error-unitary}.
\end{enumerate}

\begin{figure}[htbp]
\centering
\includegraphics[width=4in]{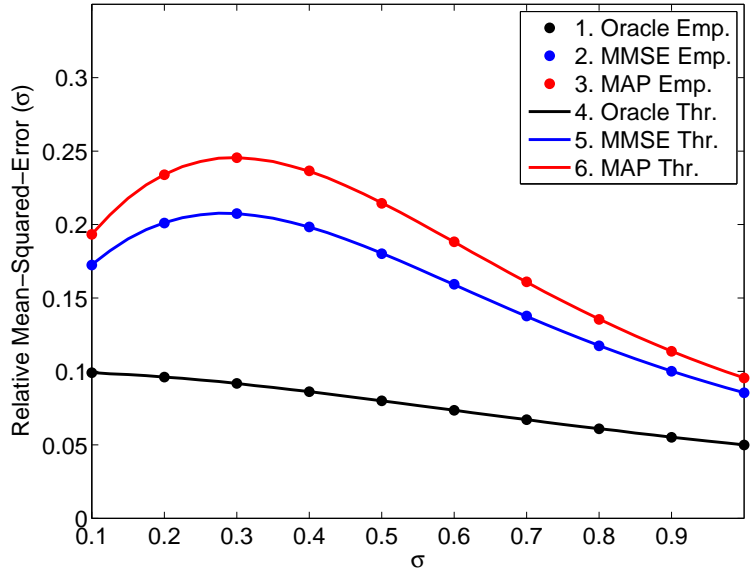}
\caption{Empirical and theoretical evaluations of the MSE as a
function of the input noise for synthetic signals ($P=0.1$,
$\sigma_x=1$, and $n=128\times128$).} \label{fig:sinDenoising}
\end{figure}

\noindent The above process is repeated for $1000$ randomly
generated signals of size $128 \times 128$, and the mean $L_2$ error
is averaged over all signals to obtain an estimate of the expected
quality of each estimator. Figure \ref{fig:sinDenoising} shows the
relative denoising effect (compared to the original noisy signal)
achieved by each estimator. The improved performance of the MMSE
estimator over the MAP is clearly seen, as well as a clear
validation of the theoretical derivations.

\subsection{Real-World Signals}

Next, we experiment with real-world signals -- images. The unitary
dictionary for this experiment is the same 2D Wavelet Transform
dictionary used in the synthetic experiment. This dictionary is
known to serve natural image content adequately (i.e., sparsify
image content). There are two main obstacles when aiming to
operate on non-synthetic signals:
\begin{enumerate}
\item The assumption that all the non-zero entries in \ve{x} share the
same variance is inadequate, and we should generalize the above
discussion to a heteroscedastic model.
\item The parameters that describe the signal model are unknown and need
to be estimated from the corrupted signal.
\end{enumerate}
\noindent Our handlng of these two issues is described in detail
in Appendix A.

It is important to note that our main goal in this experiment is
to demonstrate the power of the MMSE and the MAP estimators, and
their comparison. We do not attempt to compare these results to
state-of-the-art image denoising algorithms, as the current model
is too limited for this comparison to be fair, due to the
non-adaptiveness and the unitarity of the dictionary.

\begin{figure}
\centering
\includegraphics[width=4in]{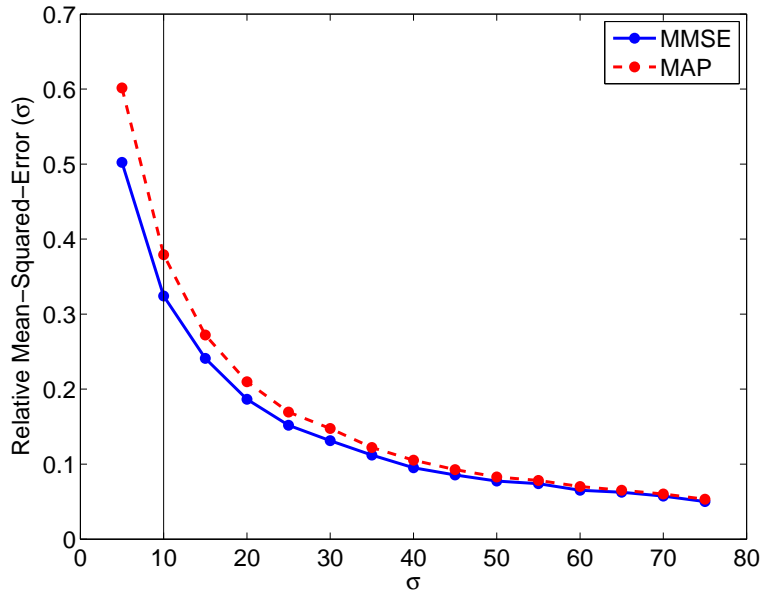}
\caption{Relative denoising achieved by the MAP and the MMSE
estimators, obtained for the image {\tt Peppers}, with varying input
noise power.} \label{fig:realDenoising}
\end{figure}

\begin{figure}
\centering \subfloat[Ground truth
image]{\includegraphics[width=2in]{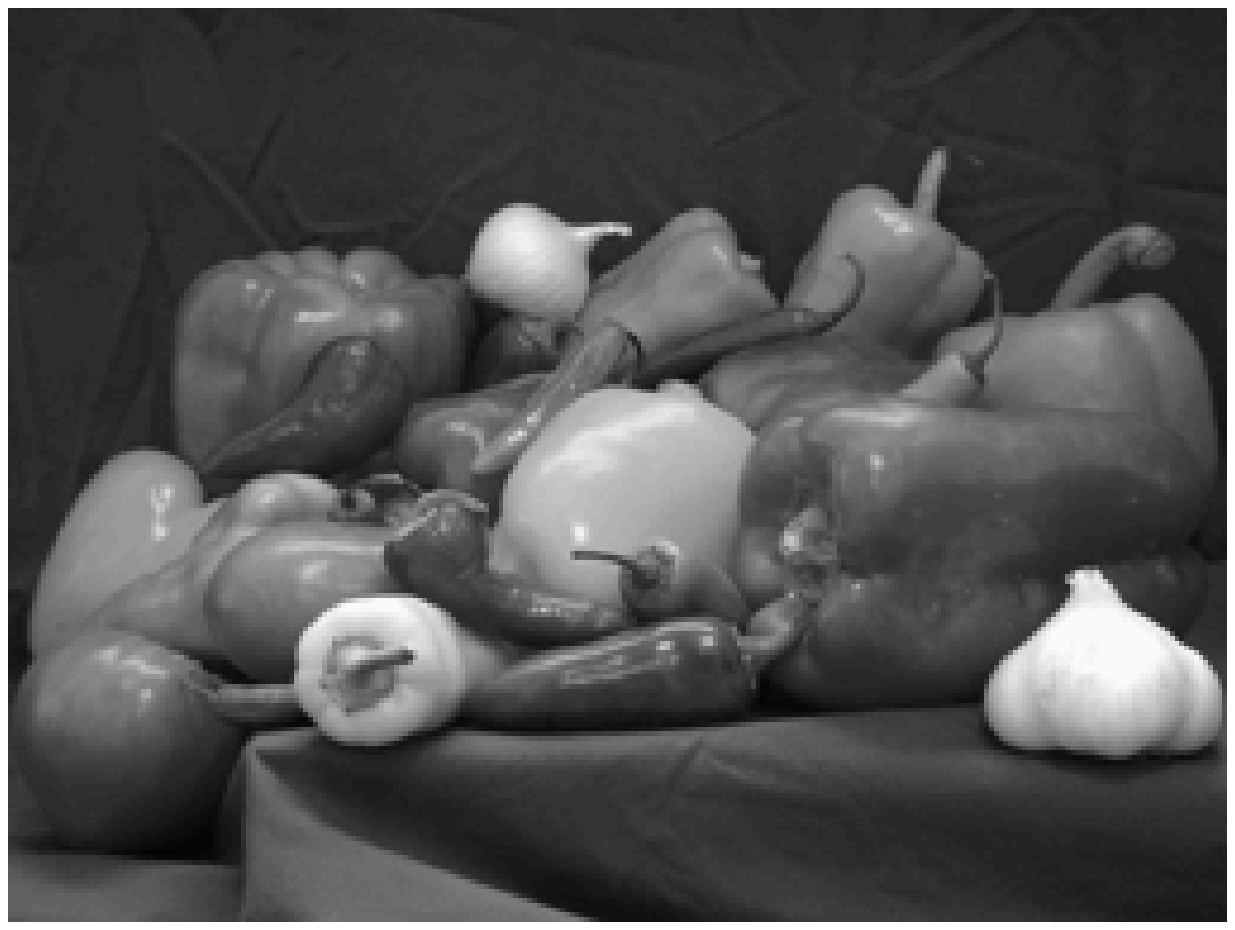}} \qquad
\subfloat[Noisy image (PSNR 28.12dB)]
{\includegraphics[width=2in]{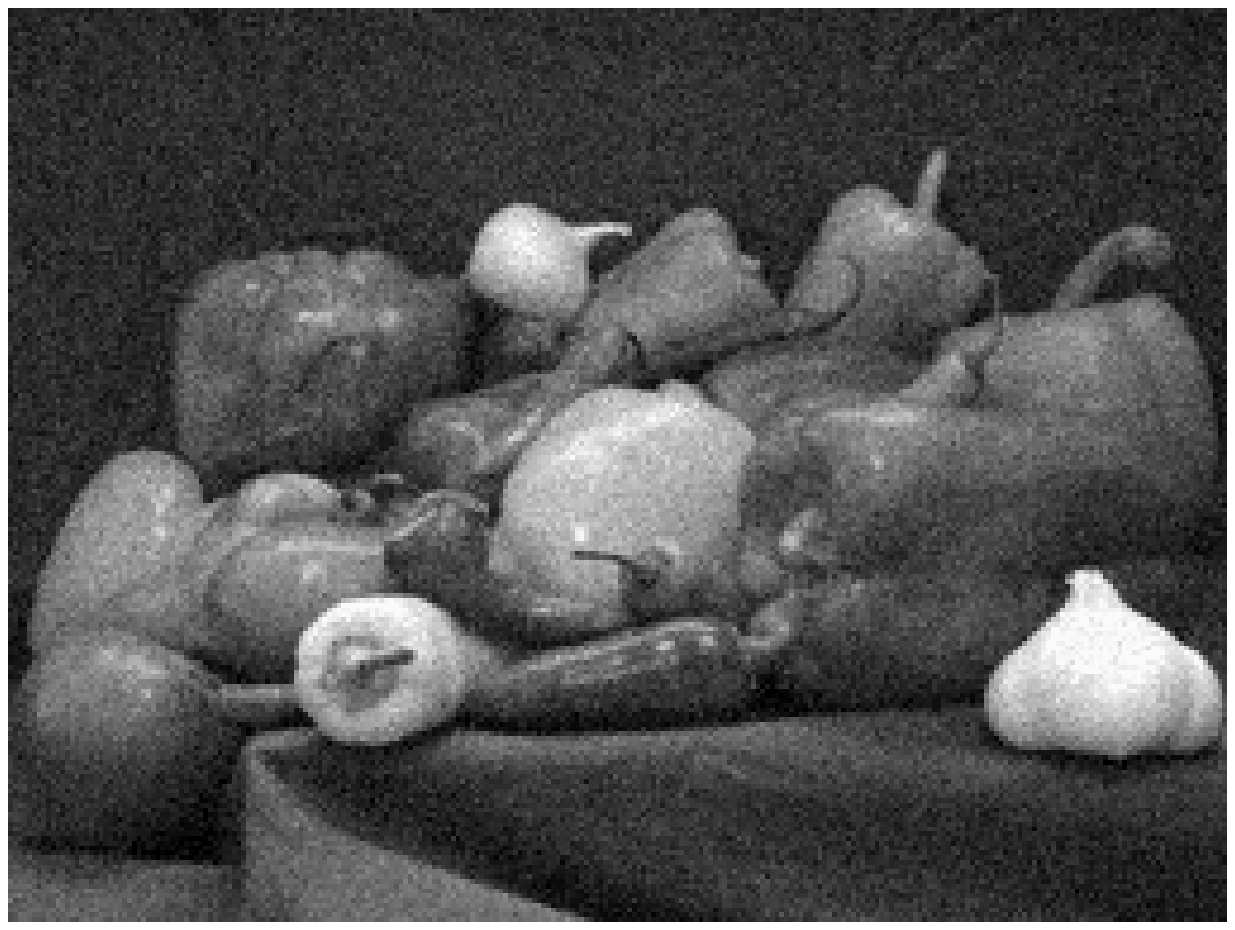}}\\
\subfloat[MAP (PSNR 32.33dB)]
{\includegraphics[width=2in]{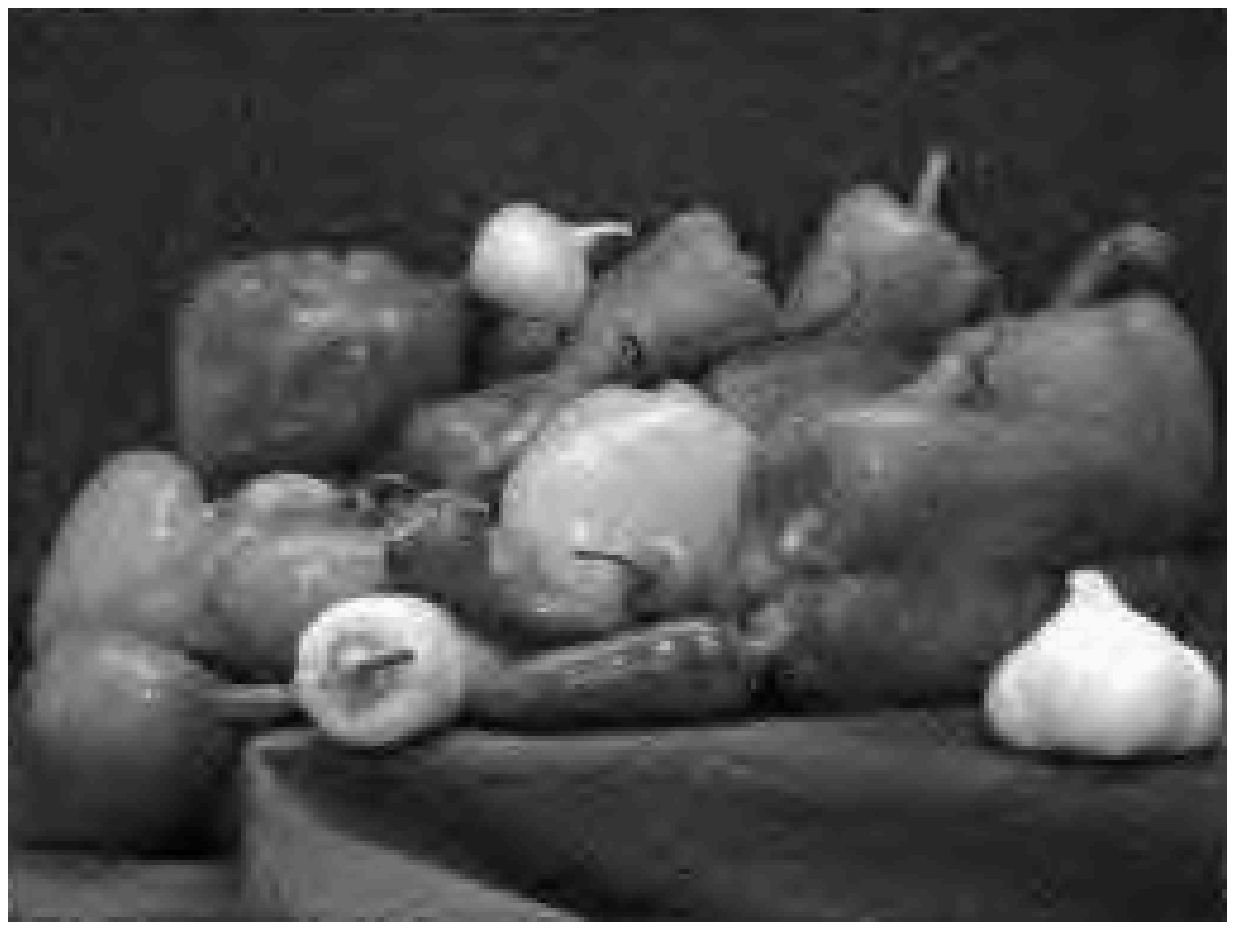}} \qquad
\subfloat[MMSE (PSNR 33.01dB)]
{\includegraphics[width=2in]{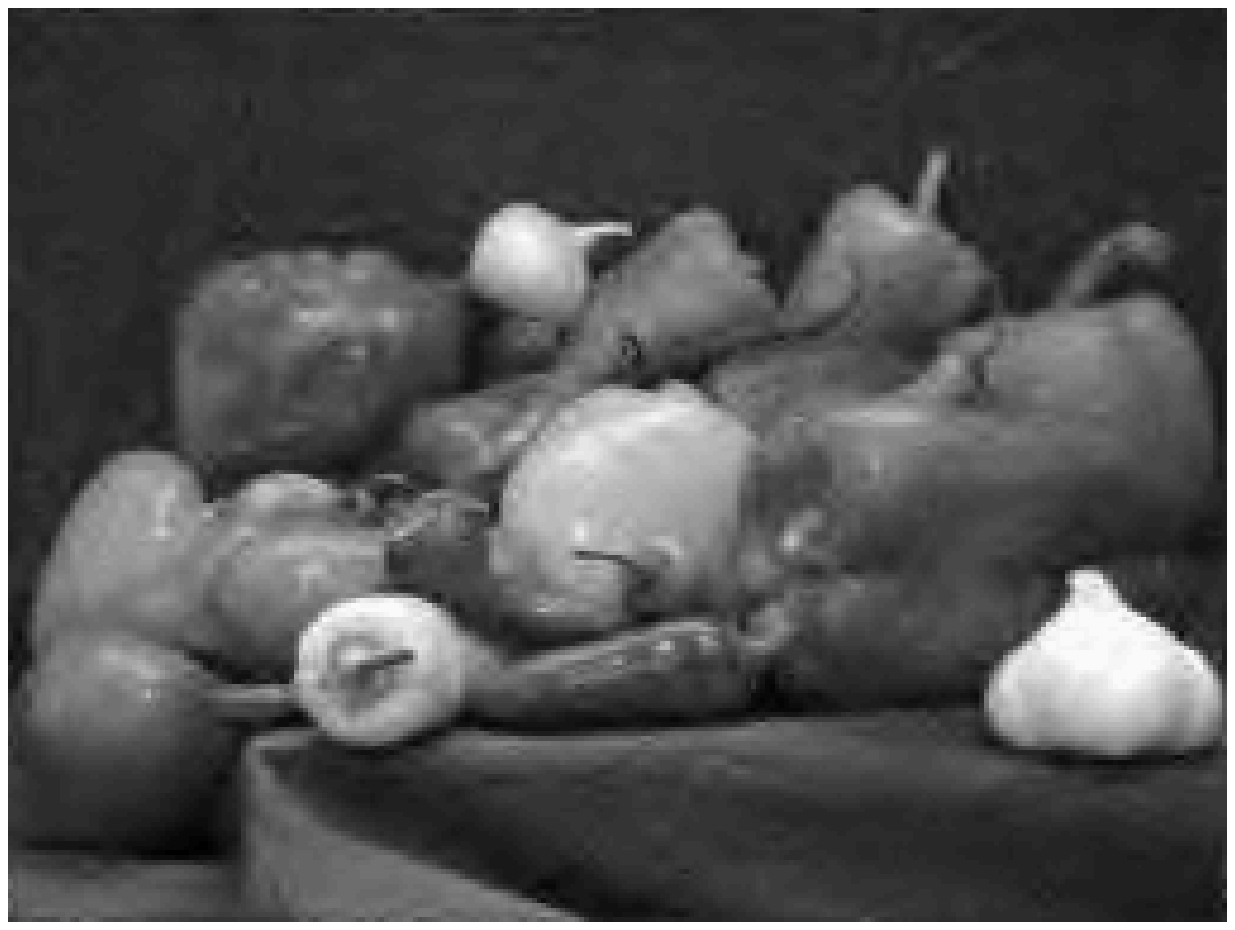}}
\caption{Visual comparison of the reconstructed image by the MAP and
MMSE estimators ($\sigma=10$).} \label{fig:realVisual}
\end{figure}

We experiment with the image {\tt Peppers} shown in Figure
\ref{fig:realVisual}. The noise levels considered are: $5,~10,~15,
\ldots~,70$, where the pixel values are in the range $[0,255]$.
The relative MSE of the cleaned image compared to the noisy one
appears in Figure \ref{fig:realDenoising}, as a function of the
input noise power. Per each $\sigma$, the parameters are
estimated, and then used within the MAP and the MMSE estimators.

\begin{figure}
\centering
\subfloat[$P_i$]{\includegraphics[width=3in]{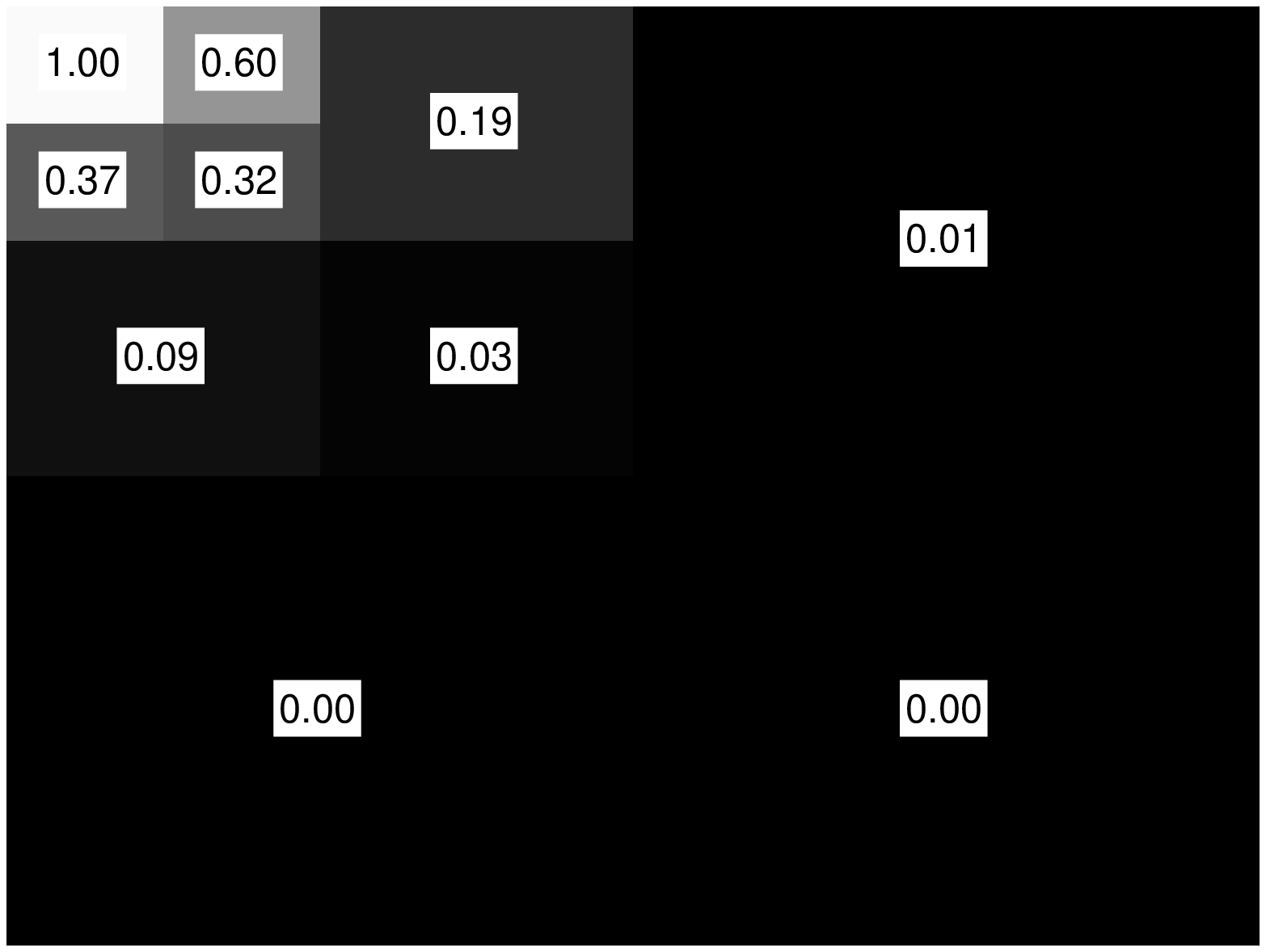}}
\subfloat[$\sigma_i$]{\includegraphics[width=3in]{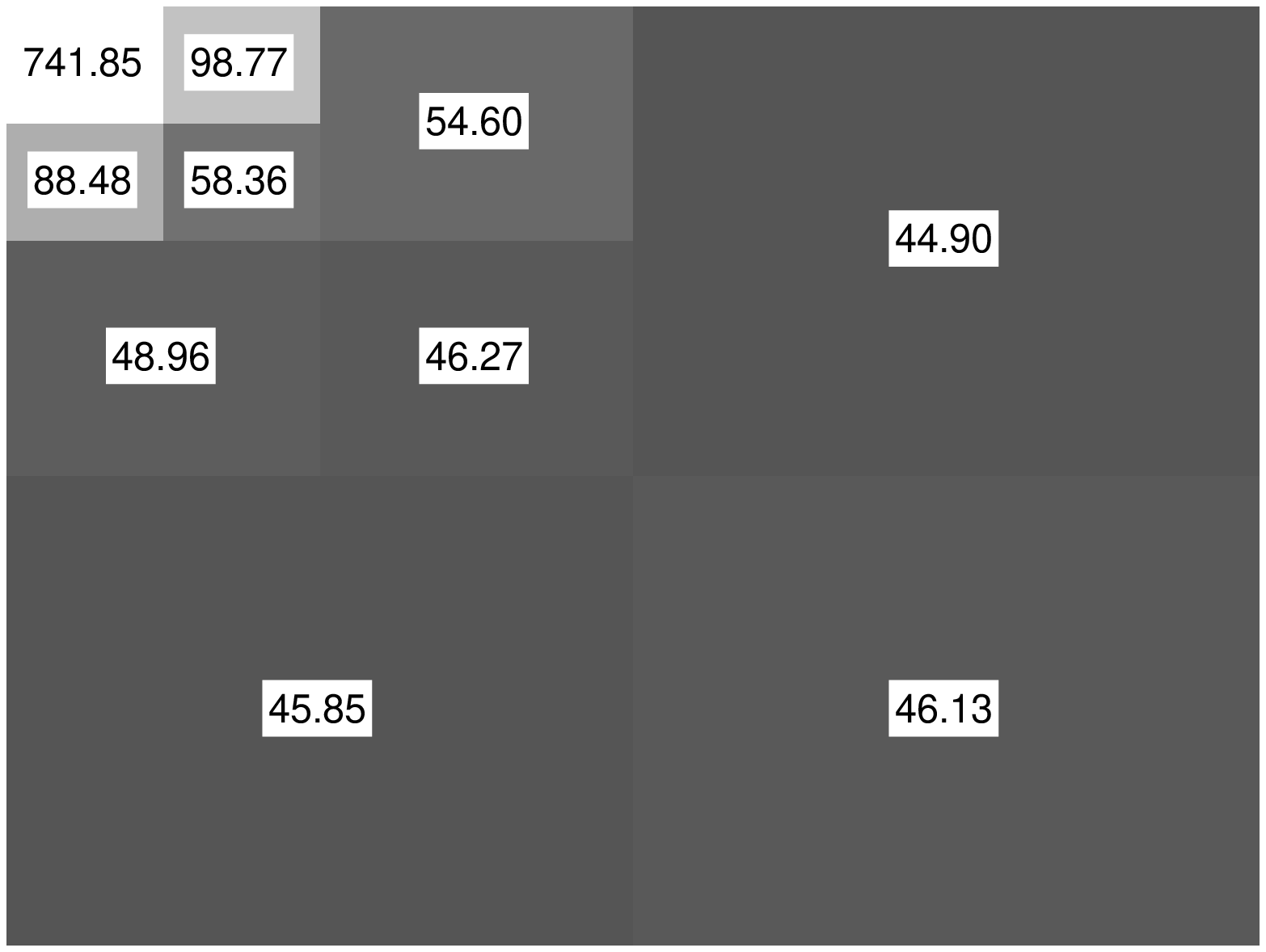}}
\caption{Estimated parameters for the ``Peppers'' image with noise
$\sigma=10$.} \label{fig:realParameters}
\end{figure}

Clearly, the MMSE outperforms the MAP for all the noise levels,
the gap being bigger for high SNR levels. Nevertheless, it is also
evident from this graph that the difference between the two is
relatively small. Figure \ref{fig:realParameters} shows the
estimated parameters learnt from the noisy frame for each band,
and the values of these parameter may provide an explanation for
this phenomenon.

\begin{figure}
\centering
\subfloat[MMSE]{\includegraphics[width=3in]{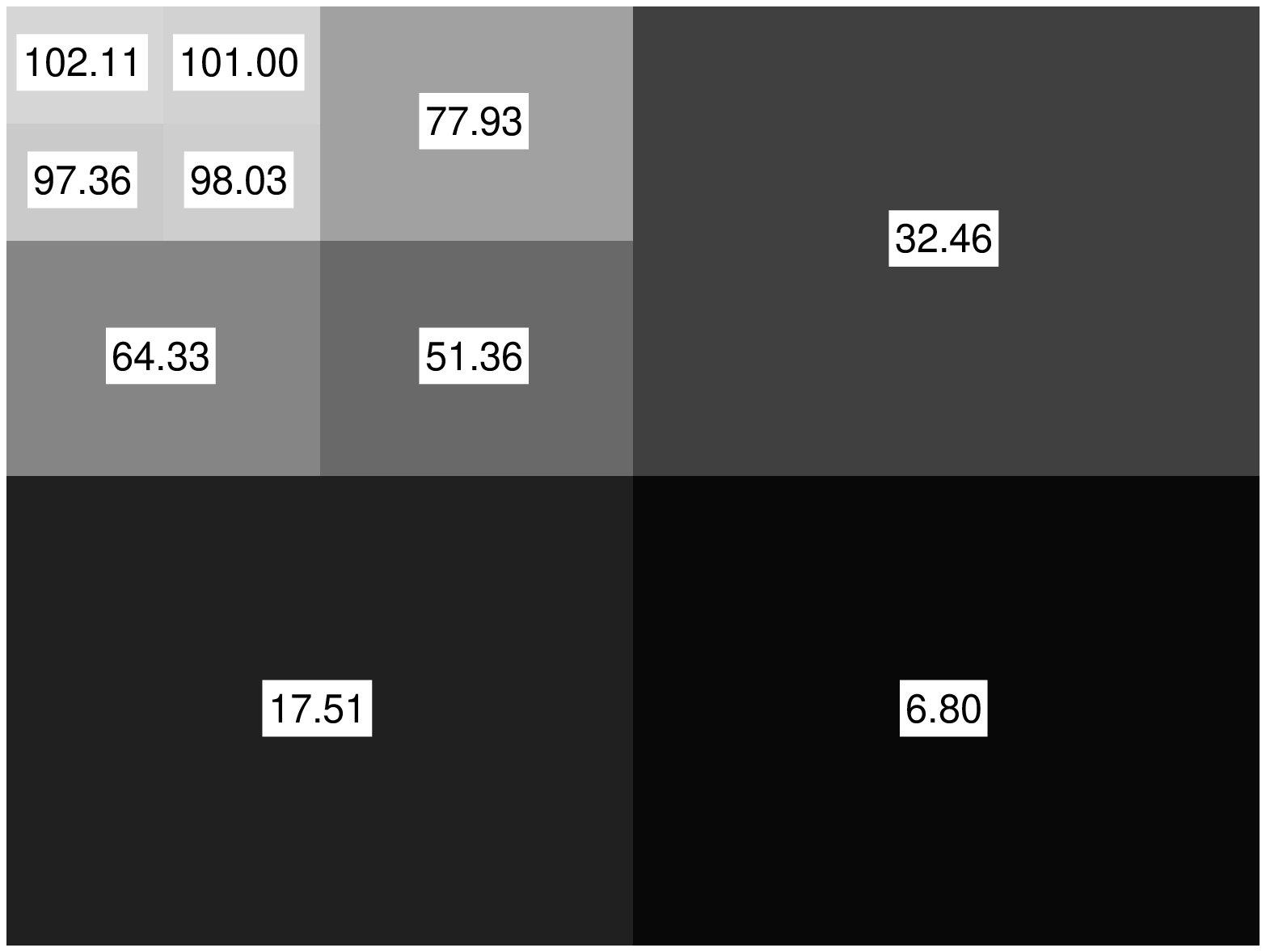}}
\subfloat[MAP]{\includegraphics[width=3in]{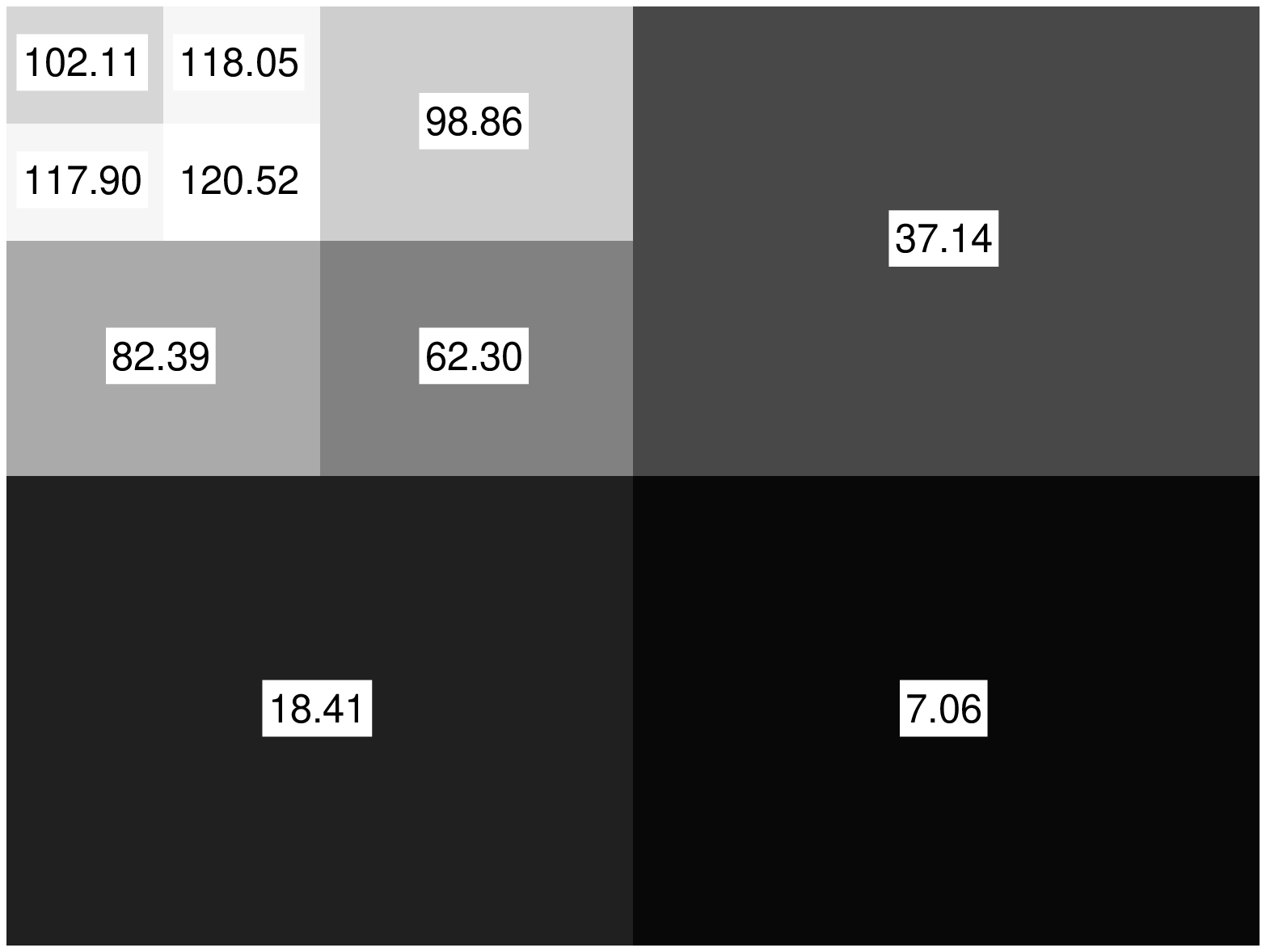}}
\caption{The error per band for the MMSE and the MAP estimators.}
\label{fig:ErrorBands}
\end{figure}

As we have observed in the previous section, the gap between the
MMSE and the MAP is expected to be negligible if $P_k$ are nearly
zeros or ones. This means that among the $10$ bands in the wavelet
transform, the three high-resolution and the single low-resolution
bands are expected to give the same performance for both
estimators. This suggests that the difference between the MAP and
the MMSE is only due to the image energy that resides in the $6$
middle-bands. Figure \ref{fig:ErrorBands} shows the actual errors
per band, as obtained by the MMSE and the MAP, and indeed, as
expected, the difference in these errors exists mostly in the $6$
middle bands.

Finally, a visual comparison of the results of the different
estimators is presented in Figure \ref{fig:realVisual} for the image
{\tt Peppers}, to which white Gaussian noise with $\sigma = 10$ is
added. As expected, the MMSE result shows a small visual improvement
over the MAP.

\section{Summary and Conclusions}
\label{sec:Summary}

In this work we have studied a  model where each atom has a given
probability to be part of the support. This model assumes that all
the supports are possible, thus avoiding assumptions on the
(generally unknown) support size. We study MAP and MMSE estimators
for the model with a general dictionary, including an overview of
their performance. Then, we focus on unitary dictionaries, for
which both estimators have simple and accurate closed formulas for
their computation. After developing the closed-form MAP and MMSE
estimators, it is shown how can they be interpreted in terms of
shrinkage. We describe the relation of the MAP and MMSE estimators
in this model to existing models appearing in the literature. This
development is extended by looking at the theoretical performance
of the estimators. Here, analytical bounds on the worst-case
denoising performance is shown. Finally, synthetic and real-world
experiments show the performance of the estimators, and the clear
advantage of MMSE estimator over MAP estimator.


\section*{Appendix A -- Handling Images}
\setcounter{section}{1} \setcounter{equation}{0}
\renewcommand{\theequation}{A-\arabic{equation}}
\renewcommand{\thesection}{\Alph{section}}

As mentioned in Section \ref{sec:Experiments}, in order to handle
a given noisy image, we should extend the model to allow for
distinct variances for the different atoms, and we should also
estimate the model parameters from the image. This appendix
describes these two tasks.

\subsection{Extension to Heteroscedastic Model}

In the derivations in this paper we have assumed that all the
non-zero entries in $\ve{x}$ have the same variance. As this is
rarely the case for natural images, we treat now a more general
problem, where this variance is atom-dependent. Such a model is
known as heteroscedastic. Our goal is to show that most of the
results remain of similar form, with modest changes. Thus, we shall
keep the discussion in the section brief, and only state the main
results.

We change the covariance matrix in Equation \eqref{eq:distXgivenS}
to be a more general diagonal matrix $\mats{V}{S}$, given by
\begin{equation}
\mats{V}{S}= \diag\left( \sigma_{\supp{S}_1}^2, \dots,
\sigma_{\supp{S}_k}^2 \right), \label{eq:matrixV}
\end{equation}
where $k = \suppsize{S}$. For the general estimators developed in
section \ref{sec:General}, the changes due to this generalization
are all absorbed in the matrices $\mats{Q}{S}$ and $\mats{C}{S}$,
becoming
\begin{eqnarray*}
\mats{C}{S} & = & \mat{D}_{\supp{S}}\mats{V}{S}\mat{D}^T_{\supp{S}} + \sigma^2\mat{I}_n, \\
\label{eq:matrixC-Hetero} \mats{Q}{S} & = & \mats{V}{S}^{-1} +
\frac{1}{\sigma^2}\mats{D}{S}^T\mats{D}{S},
\label{eq:matrixQ-Hetero}
\end{eqnarray*}
and the relation between them in Equation \eqref{eq:invMatrixC} is
still valid.

Moving to the unitary case, the matrix $\mats{Q}{S}$ is a diagonal
matrix of the form
\begin{equation}
 \mats{Q}{S} = \diag \left(\frac{\sigma_{\supp{S}_1}^2+\sigma^2}
 {\sigma_{\supp{S}_1}^2\sigma^2} ,\dots,\frac{\sigma_{\supp{S}_k}^2+
 \sigma^2}{\sigma_{\supp{S}_k}^2\sigma^2} \right).
 \label{eq:Qs-Unitary}
\end{equation}
Its inversion, $\mats{Q}{S}^{-1}$, can easily be calculated, and
the oracle solution becomes
\begin{equation}
\ve{\hat{x}}^{Oracle} = \diag\left( c_{\supp{S}_1}^2 , \dots ,
c_{\supp{S}_k}^2 \right) \cdot  \ves{\beta}{S},
\label{eq:Xoracle-Unitary-Hetero}
\end{equation}
where $c_{\supp{S}_i}^2 =
\sigma_{\supp{S}_i}^2/(\sigma_{\supp{S}_i}^2+\sigma^2)$. The support
of the MAP estimator is given by
\begin{equation}
\supp{S^{MAP}} = \arg\max_{\supp{S}\in\Omega}
\prod_{i\in\supp{S}}\sqrt{1-c^2_i} \cdot \frac{P_i}{1-P_i} \cdot
\exp\left\{\frac{c_i^2}{2\sigma^2}\ve{\beta}_i^2\right\}
\prod_{j\notin\supp{S}}1. \label{eq:MAP-Unitary-Hetero}
\end{equation}
Lastly, the unitary MMSE estimate presented in Equation
\eqref{eq:MMSE-Unitary-Closeform} becomes
\begin{equation}
\ve{\hat{x}}^{MMSE}  = \sum_{k=1}^n c_k^2 \frac{q_k}{1+q_k} \beta_k
\ve{e}_k, \label{eq:xMMSE-Unitary-Hetero}
\end{equation}
where $q_k = \frac{P_k}{1-P_k} \sqrt{1-c_k^2} \exp\left\{
\frac{c_k^2}{2\sigma^2} \beta_k^2 \right\} $.

\subsection{Parameter Estimation}

The parameters of the image generation model are not known in
advance and thus they should be estimated. We shall assume that each
band in the wavelet transform is characterized by a pair of
parameters $\sigma_{i},P_{i}$, and there are $r$ such bands overall
($10$ in the experiment reported in Section \ref{sec:Experiments}).
We propose to estimate these parameters directly from the noisy
image, by performing the following optimization task:
\begin{eqnarray}
\arg\max_{\{P_i,\sigma_i\}_{i=1}^r} P\left(\ve{y} \left|
\{P_i,\sigma_i\}_{i=1}^r  \right. \right).
\end{eqnarray}
Marginalization of this likelihood term with respect to the support
of the image in the wavelet domain reads
\begin{eqnarray}
P\left(\ve{y} \left| \{P_i,\sigma_i\}_{i=1}^r \right. \right) =
\sum_{\supp{S}} P\left(\ve{y} \left|
\supp{S},\{P_i,\sigma_i\}_{i=1}^r \right. \right)\cdot P\left(
\supp{S} \left| \{P_i,\sigma_i\}_{i=1}^r \right.\right).
\end{eqnarray}
As maximization of this summation may be computationally
difficult, we turn to approximate it by considering only one item
-- the dominant one within this sum. Thus, we propose to solve
\begin{eqnarray}\label{eq:ML}
\arg \max_{\{P_i,\sigma_i\}_{i=1}^r} &~& P\left(\ve{y} \left|
\{P_i,\sigma_i\}_{i=1}^r \right. \right) \\ \nonumber & & \approx
\arg\max_{\{P_i,\sigma_i\}_{i=1}^r,\supp{S}} P\left(\ve{y} \left|
\supp{S},\{P_i,\sigma_i\}_{i=1}^r \right. \right) \cdot  P\left(
\supp{S} \left| \{P_i,\sigma_i\}_{i=1}^r \right.\right)\cdot P\left(
\supp{S}\right),
\end{eqnarray}
where we maximize with respect to the support as well. Note that
we have introduced a prior on the support size, $P\left(
\supp{S}\right)$. We shall use the form
\begin{equation*}
P\left( \supp{S}\right)= \prod_{i=1}^r \exp\left\{ -\lambda_i
|\supp{S}_i| \right\} \, ,
\end{equation*}
with $\supp{S}_i$ the support in the $i$-th band. This prior
controls the support sparsity in each band, and as we show next,
it stabilizes the estimation procedure. The values $\lambda_i$ are
set to be high for low-frequency bands, and decrease for the higher
frequency bands.

We use the model definitions in Section \ref{sec:Model} in order to
develop an expression that depends only on the parameters of the $r$
bands. Starting with $P\left( \supp{S} \left|
\{P_i,\sigma_i\}_{i=1}^r \right.\right)$, we get
\begin{equation}
P\left( \supp{S} \left| \{P_i,\sigma_i\}_{i=1}^r \right.\right) =
\prod_{i=1}^{r}P_{i}^{|\supp{S}_{i}|}\left(1-P_{i}\right)^{n_{i}-|\supp{S}_{i}|},
\label{eq:SgivenTheta}
\end{equation}
where $n_{i}$ is the size of the $i$-th band. Using the fact that
the wavelet dictionary is unitary and exploiting Equation
\eqref{eq:Qs-Unitary}, we have
\begin{equation}
\det\left(\mats{C}{S}\right) =
\left(\frac{\sigma_{i}^{2}+\sigma^{2}}{\sigma^{2}}\right)^{|\supp{S}|}\sigma^{2n}
=\sigma^{2n}\prod_{i\in\supp{S}}\frac{\sigma_{i}^{2}+\sigma^{2}}{\sigma^{2}}
=\sigma^{2n}\prod_{i=1}^{r}\left(\frac{\sigma_{i}^{2}+\sigma^{2}}{\sigma^{2}}
\right)^{|\supp{S}_{i}|}. \label{eq:detCs-Unitary}
\end{equation}
Plugging Equation \eqref{eq:distYgivenS} and the above expressions
into \eqref{eq:ML}, the parameters estimation task becomes
\begin{equation*}
\arg\max_{\supp{S},\{P_i,\sigma_i\}_{i=1}^r} \prod_{i=1}^{r}
\left(\frac{\sigma_{i}^{2}+\sigma^{2}}{\sigma^{2}}\right)^{-\frac{|\supp{S}_{i}|}{2}}
P_{i}^{|\supp{S}_{i}|}\left(1-P_{i}\right)^{n_{i}-|\supp{S}_{i}|}\exp\left\{
\frac{1}{2\sigma^{2}}\frac{\sigma_{i}^{2}}{\sigma^{2}+\sigma_{i}^{2}}\left\Vert
\beta_{\supp{S}_{i}}\right\Vert^{2} -
\lambda_{i}|\supp{S}_i|\right\}. \label{eq:ML-4}
\end{equation*}
Two important features of this expression deserve our attention:
First, rather than seeking the support $\supp{S}$, this expression
reveals that all we need are the cardinalities $|\supp{S}|$ within
each band. Second, this expression is separable with respect to the
$r$ bands, implying that we can estimate $P_i,\sigma_i$ for the
$i$-th band by solving
\begin{equation*}
\arg\max_{|\supp{S}_i|,P_i,\sigma_i}
\left(\frac{\sigma_{i}^{2}+\sigma^{2}}{\sigma^{2}}\right)^{-\frac{|\supp{S}_{i}|}{2}}
P_{i}^{|\supp{S}_{i}|}\left(1-P_{i}\right)^{n_{i}-|\supp{S}_{i}|}\exp\left\{
\frac{1}{2\sigma^{2}}\frac{\sigma_{i}^{2}}{\sigma^{2}+\sigma_{i}^{2}}\left\Vert
\beta_{\supp{S}_{i}}\right\Vert^{2} -
\lambda_{i}|\supp{S}_i|\right\}. \label{eq:ML-4a}
\end{equation*}
Taking the log of the above expression, we obtain an alternative
function to maximize,
\begin{eqnarray}
f\left(|\supp{S}_i|,P_i,\sigma_i \right) = & &
-\frac{|\supp{S}_{i}|}{2}\log\left(\frac{\sigma_{i}^{2}+\sigma^{2}}{\sigma^{2}}\right)
+|\supp{S}_{i}| \log P_{i} \nonumber \\
& & +\left(n_{i}-|\supp{S}_{i}|\right) \log \left(1-P_{i}\right) +
\frac{1}{2\sigma^{2}}
\frac{\sigma_{i}^{2}}{\sigma^{2}+\sigma_{i}^{2}} \left\Vert
\beta_{\supp{S}_{i}}\right\Vert^{2} - \lambda_i|\supp{S}_i|.
\label{eq:ML-5}
\end{eqnarray}
To obtain the estimates for $\sigma_{i}$ and $P_{i}$ we
differentiate $f$ with respect to these unknowns. The derivative
with respect to $P_i$ leads to
\begin{equation}
0=\frac{\partial f\left(|\supp{S}_i|,P_i,\sigma_i \right)}{\partial
P_{i}} =
\frac{|\supp{S}_{i}|}{P_{i}}-\frac{\left(n_{i}-|\supp{S}_{i}|\right)}{1-P_{i}}
~~~\Longrightarrow~~~
P_{i}=\frac{|\supp{S}_{i}|}{n_{i}}.
\end{equation}
Similarly, the derivative with respect to $\sigma_i$ gives
\begin{equation}
0=\frac{\partial f\left(|\supp{S}_i|,P_i,\sigma_i
\right)}{\partial\sigma_{i}} =
-|\supp{S}_{i}|\frac{\sigma_{i}}{\sigma_{i}^{2}+\sigma^{2}}+
\frac{\sigma_{i}}{\left(\sigma^{2}+\sigma_{i}^{2}\right)^{2}}\left\Vert
\beta_{\supp{S}_{i}}\right\Vert ^{2}~~~\Longrightarrow~~~
\sigma_{i}^{2}=\frac{\left\Vert \beta_{\supp{S}_{i}}\right\Vert^{2}}
{|\supp{S}_{i}|}-\sigma^{2}.
\end{equation}

The last step in this estimation process is to discover the
cardinality $|\supp{S}_i|$. Returning to the expression to be
maximized in Equation \eqref{eq:ML-5}, we can plug in the
solutions obtained for $P_i$ and $\sigma_i$, both being functions
of $|\supp{S}_i|$. The overall expression is thus a function of
the scalar $|\supp{S}_i|$, and the maximizer value can be found by
a simple sweep of this unknown in the range $[0,n_i]$. We should
note that for every value tested, we should also update the vector
$\beta_{S}$ to include only non-zero elements of $\supp{S}_i$.
Since we are maximizing $f\left(|\supp{S}_i|,P_i,\sigma_i
\right)$, we should choose the largest entries (in absolute value)
within this vector. After this exhaustive process is done, we pick
the support size and the respective calculated parameters that
maximize the optimization task \eqref{eq:ML-5}.


\end{document}